\documentclass[runningheads]{llncs}
\usepackage{graphicx}
\usepackage[utf8]{inputenc}

\usepackage{hyperref}
\usepackage[misc]{ifsym}
\usepackage{tikz}
\usepackage{url}
\usepackage{amsmath,amssymb,amsfonts}     
\usetikzlibrary{positioning,shapes.geometric,automata,arrows,matrix,backgrounds}

\usepackage{siunitx}

\usepackage{pgfplots}
\usepackage{paralist}
\pgfplotsset{compat=1.9}
\usepgfplotslibrary{statistics}

\usepackage[acronym]{glossaries}
\usepackage{wrapfig}

\usepackage[caption=false]{subfig}
\usepackage{algorithm}
\usepackage[noend]{algpseudocode}

\usepackage{multirow}
 
\usepackage{enumitem}

\usepackage{todonotes}
\usepackage{xspace}
\usepackage{blindtext}

\usepackage[british]{babel}

\newacronym[plural=DFAs,firstplural=deterministic finite automata (DFAs)]{DFA}{DFA}{deterministic finite automaton}
\newacronym[firstplural=Markov decision processes (MDPs)]{MDP}{MDP}{Markov decision process}
\newacronym[firstplural=systems under test (SUTs)]{SUT}{SUT}{system under test}
\newacronym{DTMC}{DTMC}{discrete time Markov chain}
\newacronym{SMC}{SMC}{statistical model-checking}

\DeclareMathOperator*{\argmax}{\arg\!\max}

\newacronym{LTL}{LTL}{linear temporal logic}
\newacronym{CTL}{CTL}{computation tree logic}

\newcommand{\Dist}{\mathit{Dist}}
\newcommand{\coinFlip}{\mathit{coinFlip}}
\newcommand{\randSel}{\mathit{randSel}}

\algrenewcommand\Return{\State \algorithmicreturn{} }%

\newcommand{\ALERGIA}{\textsc{Alergia}\xspace}
\newcommand{\IOALERGIA}{\textsc{IOAlergia}\xspace}
\newcommand{\tempest}{\textsc{Tempest}\xspace}

\newcommand{\AALpy}{\textsc{AALpy}\xspace}

\definecolor{applegreen}{rgb}{0.55, 0.71, 0.0}
\definecolor{armygreen}{rgb}{0.29, 0.33, 0.13}

\newcommand{\reset}{\mathbf{reset}}
\newcommand{\step}{\mathbf{step}}

\newcommand\shieldsym[1][]{%
\raisebox{-1.5pt}{\tikzset{
    shield width/.store in=\shieldwidth,
    shield width=1.5ex,
    shield height/.store in=\shieldheight,
    shield height=1.75ex
}%
\tikz [baseline,#1] \draw (0,\shieldheight) -- (0,\shieldwidth/2) arc [radius=\shieldwidth/2, start angle=-180, end angle=0] -- (\shieldwidth,\shieldheight) -- cycle;%
}}
\newcommand{\shielded}[2][]{\ensuremath{#2_{\text{\shieldsym}#1}}}

\newcommand{\mdp}{\mathcal{M}}

\newcommand{\trans}{\mathcal{P}}
\newcommand{\states}{\mathcal{S}}
\newcommand{\Act}{\mathcal{A}}
\newcommand{\rewards}{\mathcal{R}}
\newcommand{\lab}{L}
\newcommand{\policy}{\sigma}
\newcommand{\policies}{\Sigma}

\newcommand{\shield}{\pi}

\newcommand{\maxeplen}{\ensuremath{t_{max}}}

\newcommand{\e}{\ensuremath{\mathbb{E}}}

\newcommand{\mcresmin}[2]{\ensuremath{\eta^{\min}_{#1,#2}}}
\newcommand{\mcresmax}[2]{\ensuremath{\eta^{\max}_{#1,#2}}}

\newcommand{\pit}[2]{
  \fill[gray] (#1,#2) rectangle (#1+1,#2+1);
  \draw (#1+0.5,#2+0.5) node{\Large$\mathit{pit}$};
}
\newcommand{\wall}[2]{
  \fill[black] (#1,#2) rectangle (#1+1,#2+1);
}

\newcommand{\celllab}[4]{
  \def\LABEL{#4}
  \fill[#3] (#1,#2) rectangle (#1+1,#2+1);
  \draw (#1+0.5,#2+0.5) node{\large$\mathit{\LABEL}$};
}
\newcommand{\slipdown}[2]{
  \shade[top color=white,bottom color=gray] (#1,#2) rectangle (#1+1,#2+1);
  \draw[very thick,->] (#1+0.6,#2+0.6) -- (#1+0.3,#2+0.3);
}
\newcommand{\slipdownhard}[2]{
  \shade[top color=white,bottom color=gray] (#1,#2) rectangle (#1+1,#2+1);
  \draw[very thick,->] (#1+0.8,#2+0.8) -- (#1+0.2,#2+0.2);
}
\newcommand{\slipup}[2]{
  \shade[top color=gray,bottom color=white] (#1,#2) rectangle (#1+1,#2+1);
  \draw[very thick,->] (#1+0.6,#2+0.3) -- (#1+0.3,#2+0.6);
}

\newcommand{\slipuphard}[2]{
  \shade[top color=gray,bottom color=white] (#1,#2) rectangle (#1+1,#2+1);
  \draw[very thick,->] (#1+0.8,#2+0.2) -- (#1+0.2,#2+0.8);
}

\newcommand{\slipleft}[2]{
  \shade[top color=lightgray,bottom color=white] (#1+1,#2+1) rectangle (#1,#2);
  \draw[very thick,->] (#1+0.6,#2+0.5) -- (#1+0.4,#2+0.5);
}

\usepackage[utf8]{inputenc}
\usepackage{pgfplots}
\DeclareUnicodeCharacter{2212}{−}
\usepgfplotslibrary{groupplots,dateplot}
\usetikzlibrary{patterns,shapes.arrows}
\pgfplotsset{compat=newest}

\begin{document}
\title{Automata Learning meets Shielding}
%
%

\author{
Martin Tappler\inst{1,2}\and
Stefan Pranger\inst{1}\and
Bettina K\"onighofer\inst{1,3} \and
Edi Mu\v{s}kardin\inst{2,1} \and
Roderick Bloem\inst{1,2} \and
Kim Larsen\inst{4}}
\authorrunning{Tappler et al.}
%
\institute{
Graz University of Technology \and
TU Graz-SAL DES Lab,Silicon Austria Labs, Graz, Austria \and
Lamarr Security Research \and
Aalborg University, Aalborg, Denmark\\
\email{
martin.tappler@ist.tugraz.at,
stefan.pranger@iaik.tugraz.at,
bettina.koenighofer@lamarr.at,
edi.muskardin@silicon-austria.com,
roderick.bloem@iaik.tugraz.at,
kgl@cs.aau.dk
}
}
\maketitle              
\begin{abstract}
Safety is still one of the major research challenges in reinforcement learning (RL).
In this paper, we address the problem of how to avoid safety violations of RL agents during exploration in probabilistic and partially unknown environments. 
Our approach combines automata learning for Markov Decision Processes (MDPs) and shield synthesis in an iterative approach. Initially, the MDP representing the environment is unknown. The agent starts exploring the environment and collects traces.
From the collected traces, we passively learn MDPs that abstractly represent the 
safety-relevant aspects of the environment.
Given a learned MDP and a safety specification, we construct a shield.
For each state-action pair within a learned MDP, the shield computes exact probabilities on how likely it is that executing the action results in violating the specification from the current state within the next $k$ steps. 
After the shield is constructed, the shield is used during
runtime and blocks any actions that induce a too large risk from the agent.
The shielded agent continues to explore the environment and collects new data on the environment.
Iteratively, we use the collected data to learn new MDPs with higher accuracy,
resulting in turn in shields able to prevent more safety violations.
We implemented our approach and present a detailed case study of a Q-learning agent exploring slippery Gridworlds. In our experiments, we show that as the agent explores more and more of the environment during training, the improved learned models lead to shields that are able to prevent many safety violations.


\end{abstract}
\section{Introduction}

Nowadays systems are increasingly autonomous and make extensive use of machine learning.
The tremendous potential of autonomous, AI-based systems is contrasted by the growing concerns about their safety~\cite{DBLP:conf/uai/CorsiMF21}. Their huge complexity makes it infeasible to formally prove their correctness or to cover the entire input space of a system with test cases.
An especially challenging problem is ensuring safety during the learning process~\cite{DBLP:conf/isola/KonighoferL0B20}.
In model-free \emph{reinforcement learning} (RL)~\cite{DBLP:books/lib/SuttonB98}, an agent aims to learn a task through trial-and-error via interactions with an unknown environment. While the model-free RL approach is very general as well as scalable and has successfully been applied in various challenging application domains~\cite{kiran2021deep}, the learning agent needs to explore unsafe behavior in order to learn that it is unsafe.


\emph{Shielding}~\cite{DBLP:conf/aaai/AlshiekhBEKNT18} is a runtime enforcement technique that applies correct-by-construc\-tion methods to automatically compute shields from a given safety temporal logic specification~\cite{BK08} and a model that captures all safety-relevant dynamics of the environment. 
Shields have been categorized into post-shields and pre-shields.
Post-shields monitor the actions selected by the agent and overwrite any unsafe action with a safe one. 
Pre-shields are implemented before the agent and block, at every time step, unsafe actions from the agent (also referred to as action masking). Thus, the agent can only choose from the set of safe actions. 
In this paper, we use pre-shielding since this setting allows the agent maximal freedom in exploring the environment.

In the \emph{non-probabilistic setting}~\cite{DBLP:conf/tacas/BloemKKW15}, shields guarantee that the safety specification will never be violated, working under the assumption that a complete and faithful environmental model of the safety-relevant dynamics is available. Shielding in the \emph{probabilistic setting}~\cite{DBLP:conf/concur/0001KJSB20}, which is the standard setting in RL, assumes to have an environmental model in form of a Markov decision process (MDP) available.
Given such an MDP $\mathcal{M}$ and a safety specification $\varphi$,
the shield computes how likely it is that executing an action from the current state will result in violating $\varphi$ within a given finite horizon.
At any state $s$, an actions $a$ is called \emph{unsafe} if executing $a$ incurs a probability of violating $\varphi$ within the next $k$ steps greater than a relative threshold $\lambda$ w.r.t. the optimal safety probability possible in $s$.
The resulting shield prohibits safety violations that can be prevented by planning ahead $k$ steps into the future. 
Shielding requires a complete and accurate environmental model, but it is rarely the case that such a model is available. However some data about the environment often exists, 
for example, a RL agent collects data by exploring the environment. 

\emph{Automata learning}~\cite{DBLP:journals/cacm/Vaandrager17,DBLP:conf/dagstuhl/AichernigMMTT16,DBLP:conf/dagstuhl/HowarS16} is a well-established technique to automatically learn automata models of black-box systems from observed data.
The data used for automata learning is usually given in the form of observation traces, which are sequences of observations of the environment's state
and actions chosen by the agent.
\emph{Passive MDP learning}~\cite{DBLP:journals/corr/abs-1212-3873,DBLP:journals/ml/MaoCJNLN16} is able to learn MDP models from a multiset of sampled observation 
traces. Thus, the learned MDP depends on the given sampled traces. 

\noindent\textbf{Our approach.} In this paper, we consider the setting of RL in an initially unknown environment. The goal is to reduce safety violations during the exploration phase of the RL-agent by combining \emph{passive MDP learning} with \emph{probabilistic shielding} in an iterative approach.
Initially, the MDP representing the environment
is unknown. 
During runtime, the agent collects observation traces while exploring the environment. After having a large enough initial multiset of traces, we first transform the sequences of observed states in the traces into observations, which include only the safety-relevant information, using a suitable abstraction function. 
The abstract traces are then used to learn a first estimate of the safety-relevant MDP.
From this initial MDP and a given safety specification, we
construct an initial shield. 
After the shield is constructed,
the agent is augmented with the shield, i.e., the shield blocks unsafe actions from the agent and the agent can pick from the set of safe actions. 
The newly collected traces are added to the multiset of all traces.
After collecting a predefined number of new traces, our approach learns a new safety-relevant MDP from the multiset of all collected traces and creates a new shield.

At every iteration, the shield is built from an MDP that approaches more and more the real MDP underlying the environment modulo the abstraction to safety-relevant observations. Thus, the resulting shields are getting more informed and 
prevent the agent from entering more safety-critical situations. 

\noindent\textbf{Outline.} The rest of the paper is structured as follows.
We present the related work in Section~\ref{sec:related} and discuss the 
relevant foundations in Section~\ref{sec:prelim}.
We present our approach for safe learning via shielding and automata learning in 
Section~\ref{sec:method}. We present our experimental results in Section~\ref{sec:experiments}
and conclude in Section~\ref{sec:conclusion}.

\section{Related Work}
\label{sec:related}
We combine automata learning and probabilistic verification to create 
safety shields in our approach. Early work on such combinations
has been performed by Cobleigh et al.~\cite{DBLP:conf/tacas/CobleighGP03},
who propose to learn assumptions for compositional reasoning. 
More closely related to our work is black box checking by
Peled et al.~\cite{DBLP:journals/jalc/PeledVY02}, where
they present a technique for model checking of 
deterministic black-box systems. Learning-based testing 
by Meinke and Sindhu~\cite{DBLP:conf/tap/MeinkeS11} follows a similar approach of incremental 
learning of hypothesis models and model checking of these hypotheses.
In previous work~\cite{DBLP:journals/fmsd/AichernigT19}, we proposed a technique inspired by black box checking
for probabilistic reachability checking of stochastic black-box systems.
As in this paper, we applied \IOALERGIA~\cite{DBLP:journals/corr/abs-1212-3873,DBLP:journals/ml/MaoCJNLN16} to learn MDPs. Rather than computing safety shields from learned
MDPs, we computed policies to satisfy reachability objectives.
We also proposed a technique for $L^*$-based learning of MDP~\cite{DBLP:journals/fac/TapplerA0EL21}, which may serve as a
basis for RL and shielding. 
In this paper, we combine stochastic learning and abstraction with respect to safety-relevant features to improve RL. Nouri et al.~\cite{DBLP:conf/rv/NouriRBLB14} also apply abstraction on traces 
with respect to properties of the system under consideration in order
to learn abstract probabilistic models.
In contrast to us, they aim to improve the runtime of statistical model-checking and they learn Markov chains 
that are not controllable via inputs.


Recently, various authors have proposed combinations of automata 
learning and reinforcement learning~\cite{DBLP:conf/aaai/GaonB20,DBLP:conf/aips/0005GAMNT020,DBLP:conf/nips/IcarteWKVCM19,DBLP:conf/aaai/Furelos-BlancoL20}. By learning finite-state models, such as so-called 
reward machines, they provide additional high-level structure for RL.
This enables RL when rewards are non-Markovian, i.e., the gain depends not only
on the current state and action, but on the path taken by the agent. DeepSynth~\cite{DBLP:conf/aaai/HasanbeigJAMK21} follows a similar approach to improve
RL with sparse rewards. Related to these approaches, Muskardin et al.~\cite{DBLP:journals/corr/abs-2206-11708} propose a combination of reinforcement learning  and automata learning to handle partial observability, i.e., non-Markovian environments.

Fu and Topcu~\cite{DBLP:conf/rss/FuT14} presented an approach for a learning-based synthesis of policies for MDPs w.r.t. temporal logic specifications
that are probably approximately correct. In contrast to us, they assume the topology of the MDP 
to be known, so that only transition probabilities need to be learned.

Alshiekh et al.~\cite{DBLP:conf/aaai/AlshiekhBEKNT18} proposed shielding for RL.
Jansen et al.~\cite{DBLP:conf/concur/0001KJSB20} proposed the first method 
to compute safety shields using a bounded horizon in MDPs.
Giacobbe et al.~\cite{GiacobbeHKW21} applied the same technique on 31 Atari 2600 games.
The approach was further extended by K\"onighofer et al.~\cite{DBLP:conf/nfm/KonighoferRPTB21}. Instead of analyzing the safety of all state-action pairs ahead of time, the approach uses the time between two successive 
decisions of an agent to analyze the safety of actions on the fly.
Pranger et al.~\cite{DBLP:journals/corr/abs-2010-03842} proposed an iterative approach to shielding that updates the transition probabilities of the MDP based on observed behavior
and computes new shields in regular intervals. 
To construct our shields, we use the approach proposed by Jansen et al.~\cite{DBLP:conf/concur/0001KJSB20}. Similarly to   
Pranger et al.~\cite{DBLP:journals/corr/abs-2010-03842}, we iteratively construct new shields, but do not rely on a known topology of the MDP.

As in our approach, Waga et al.~\cite{DBLP:journals/corr/abs-2207-13446} use automata learning to dynamically construct shields during runtime. The main difference to our work is that they assume that the environment behaves deterministically, whereas we allow  probabilistic environmental behavior which is the standard assumption in reinforcement learning.

\section{Preliminaries}
\label{sec:prelim}


\paragraph*{Basics. } 
Given a set $E$, we denote by $\Dist(E)$ the set of probability distributions over $E$, i.e. for all $\mu$ in
$\Dist(E)$ we have $\mu : E \rightarrow [0,1]$ such that $\sum_{e\in E} \mu(e) = 1$.
In Section~\ref{sec:method}, we apply two randomized functions $\coinFlip$ and $\randSel$. The function
$\coinFlip$ is defined by $\mathbb{P}(\coinFlip(p) = \top) = p$ 
and $\mathbb{P}(\coinFlip(p) = \bot) = 1-p$ for  $p \in [0,1]$. The function $\randSel$ samples an element $e$ from a given set $E$
according to uniform distribution, i.e., $\forall e \in E : \mathbb{P}(\randSel(E) = e) = \frac{1}{|E|}$.

\subsection{Markov Decision Processes and Reinforcement Learning}



\begin{definition} A \textbf{Markov decision process (MDP)} is a tuple $\langle \mathcal{S},s_0,\mathcal{A}, \trans\rangle$ where $\mathcal{S}$ is a finite set of states,
 $s_0 \in \mathcal{S}$ is the initial state,
$\mathcal{A}$ is a finite set of actions,
 and
 $\trans : \mathcal{S} \times \mathcal{A} \rightarrow \Dist(\mathcal{S})$ is the probabilistic transition function.
\end{definition}
For all $s \in \states$ the available actions are $\Act(s) = \{a \in \Act\: |\: \exists s', \trans(s, a)(s') \neq 0\}$ and we assume $|\Act(s)| \geq 1$.
We associate an MDP $\mathcal{M}$  
with a \emph{reward function} $\rewards : \states \times \Act \times \states \rightarrow \mathbb{R}$.

\noindent\textbf{Traces.} A finite \emph{path} $\rho$ through an MDP is an alternating sequence 
of states and actions, i.e. 
$\rho = s_0 a_1 s_1 \cdots a_{n-1} s_{n-1} a_n s_n \in s_0 \times (\Act \times \states)^*$. The set of 
all paths of an MDP $\mathcal{M}$ is denoted by $Path_\mathcal{M}$.
We refer to a path augmented with the gained reward as a \emph{reward trace} $\tau = s_0 a_1 r_1 s_1  \cdots  s_{n-1} a_n r_n s_n$ with $r_i = \rewards(s_{i-1},a_i,s_i)$. In the remainder of the paper, we 
treat reward traces also as sequences of triples $(a_i,r_i,s_i)$
comprising an action, the gained reward, and the reached state.


\noindent\textbf{Policies.} A memoryless policy defines for every state in an MDP a probability distribution over actions. 
 Given an MDP $\mathcal{M} = \langle S,A,s_0, \trans \rangle$, a \emph{memoryless policy}
 for $\mathcal{M}$ is a function $\sigma : \states \rightarrow \Dist(\Act)$.
A \emph{memoryless deterministic policy} $\policy: \states \rightarrow \Act$ is a function over action given states. 


\noindent\textbf{Reinforcement Learning.}
An RL agent learns 
a task through trial-and-error via interactions with
an unknown environment.
The agent takes \emph{actions} and receives feedback in form
from \emph{observations} on the state of the environment and \emph{rewards}. The goal of the agent is to maximize the expected accumulated reward.

Typically, the environment is modeled as an MDP $\mathcal{M} = \langle \mathcal{S},s_0,\mathcal{A}, \trans\rangle$ with associated reward function $\rewards$. At each step $t$ of a training episode,
the agent receives an observation $s_t$. 
It then chooses an action $a_{t+1} \in \Act$. The environment then
moves to a state $s_{t+1}$ with probability $\trans(s_t, a_{t+1})(s_{t+1})$.
The reward is determined with $r_{t+1} = \rewards(s_t, a_{t+1}, s_{t+1})$.
We refer to negative rewards $r_t<0$ as \emph{punishments}.
The \emph{return} $\texttt{ret}=\Sigma^{\infty}_{t=1} \gamma^t r_t$ is the cumulative future discounted reward, where $r_t$
is the immediate
reward at time step $t$, and $\gamma \in [0, 1]$ is the discount factor that controls the influence of
future rewards.
The objective of the agent is to learn an \emph{optimal policy} $\policy^\star : \states \rightarrow \Act$ that
maximizes the expectation of the return, i.e. $\max_{\policy\in\policies} \e_\policy(\texttt{ret})$.
A training episode ends after a maximum episode length of $t_{max}$ steps. 

Q-learning is one of the most established RL algorithms.
The Q-function for policy $\policy$ is defined as the
expected discounted future reward gained by taking an action $a$ from a state $s$
and following policy $\policy$ thereafter.
Tabular Q-learning~\cite{DBLP:journals/ml/WatkinsD92} uses the experience $(s_t, a_t, r_t, s_{t+1})$ to learn the Q-function $Q^\star(s,a)$
corresponding to an optimal policy $\policy^\star(s,a)$.
The update rule is defined as 
\begin{equation*}
    Q(s_i, a_{i+1}) \gets (1 - \alpha) \cdot Q(s_i, a_{i+1}) + 
    \alpha (r_{i+1} + \gamma \cdot \max_{a \in A}(Q(s_{i+1},a))),
\end{equation*}
where $\alpha$ is the learning rate and $\gamma$ is the discount factor.

\begin{definition}
\label{def:det_mdps}
 A \textbf{deterministic labeled MDP}
 $\mathcal{M}_L = \langle \states,s_0,\Act, \trans, \lab \rangle$ is an MDP with a labeling function $\lab : \states \rightarrow O$ mapping states to observations from a finite set $O$.
 The transition function $\trans$ must satisfy the following determinism property: 
$\forall s \in \states, \forall a \in \Act : \delta(s,a)(s') > 0 \land \delta(s,a)(s'') > 0$ implies
$s' = s''$ or $\lab(s') \neq \lab(s'')$. 
\end{definition}

In this paper, we use passive automata learning to compute abstract MDPs 
of the environment in the form of deterministic labeled MDPs.
These MDPs represent safety-related information only and will not be used for RL but for shielding. 
Therefore, there is no need to use rewards in combination with
deterministic labeled MDPs. 


Given a path $\rho$ in a deterministic labeled MDP $\mathcal{M}_L$.
Applying the labeling function on all states of a path $\rho$ 
results in a so called \emph{observation trace} $L(\rho) = L(s_0) a_1 L(s_1) \cdots a_{n-1} L(s_{n-1}) a_n L(s_n)$.
Note that due to determinism, an observation trace  $L(\rho)$ uniquely
identifies the corresponding path $\rho$.

\subsection{Learning of MDPs}

We learn deterministic labelled \glspl*{MDP} via  \IOALERGIA~\cite{DBLP:journals/corr/abs-1212-3873,DBLP:journals/ml/MaoCJNLN16}, which is an adaptation of \ALERGIA{}~\cite{Carrasco_Oncina_1994}.
\IOALERGIA takes a multiset $\mathcal{T}_o$ of observation traces as input and first constructs a tree representing the observation traces, by merging 
common prefixes. The tree has edges labeled with actions and nodes
that are labeled with observations. Each edge corresponds to 
a trace prefix with the label sequence that is visited by traversing the 
tree from the root to the edge. 
Additionally, edges are associated
with frequencies that denote how many traces in $\mathcal{T}_o$ have 
the trace corresponding to an edge as a prefix. Normalizing these 
frequencies would already yield tree-shaped MDP.

For generalization, the tree is transformed into an MDP with cycles through
an iterated merging of nodes. Two nodes are merged if they are compatible, i.e., 
their future behavior is sufficiently similar. For this purpose, we check
whether the observations in the sub-trees originating in the nodes 
are not statistically different. The parameter $\epsilon_\mathrm{\ALERGIA{}}$
controls the significance level of the applied statistical tests. 
If a node is not compatible with any other potential node, it is 
promoted to an MDP state. Once all potential pairs of nodes have been 
checked, the final deterministic labeled MDP is created by normalizing 
the frequencies on the edges to yield probability distributions for the
transition function $\trans$. In this paper, we refer to this construction
as MDP learning and we denote calls to \IOALERGIA by 
$\shielded{\mathcal{M}} = 
\IOALERGIA{}(\mathcal{T}_o,\epsilon_\mathrm{\ALERGIA{}})$, where 
$\shielded{\mathcal{M}}$ is the deterministic labeled MDP learned from the multiset of observation traces $\mathcal{T}_o$.


\subsection{Shielding in MDPs}

\textbf{Specifications and Model Checking}. We consider specifications given in the safety fragment of linear temporal logic (LTL)~\cite{BK08}.
For an MDP $\mdp$ and a safety specification $\varphi$, probabilistic model checking  employs linear programming or value iteration to compute the probabilities of all states and actions of the $\mdp$ to
satisfy an $\varphi$.
Specifically, the probabilities $\mcresmax{\varphi}{\mdp} \colon \mathcal{S} \times \mathbb{N} \rightarrow [0,1]$ or $\mcresmin{\varphi}{\mdp} \colon \mathcal{S} \times \mathbb{N} \rightarrow [0,1]$
give for all states the maximal (or minimal) probability over all possible policies to satisfy $\varphi$, within a given number of steps.
For instance, a safety property $\varphi = \mathbf{G}(\lnot \mathcal{S}_{unsafe})$ could encode  that a set of unsafe states
${S}_{unsafe}\in\states$ must not be entered. 
Then $\mcresmax{\varphi}{\mdp}(s,h)$ is the maximal probability to not visit ${S}_{unsafe}$ 
from state $s\in \mathcal{S}$ in the next $h$ steps.

\noindent\textbf{Shield Construction.}
Given an MDP $\mdp$, a safety specification $\varphi$, and a finite horizon $h$,
the task of the shield is to limit the probability to violate the safety specification $\varphi$ within the next $h$ steps.

For any state $s\in \states$ and 
action $a \in \Act(s)$,
the \emph{safety-value} $val_{\varphi,\mathcal{M}}(s,a,h)$ is computed which gives
the maximal probability to stay safe from $s$ after executing $a$, i.e.,
$$val_{\varphi,\mathcal{M}}(s,a,h)=\mcresmax{\varphi}{\mdp}(\trans(s,a),h-1).$$
The \emph{optimal safety-value} $optval_{\varphi,\mathcal{M}}(s)$
of $s$ is the maximal safety value of any action $a$ in state $s$ within the next $h-1$ steps, i.e.,
$$optval_{\varphi,\mathcal{M}}(s,h) = \max_{a\in\Act(s)} val_{\varphi,\mathcal{M}}(s,a,h)=\mcresmax{\varphi}{\mdp}(\trans(s,a),h-1).$$

An action $a$ in $s$ is \emph{unsafe} if the
safety value of $a$ is lower
than the optimal safety-value by some threshold $\lambda_{sh}$, i.e., $\text{an action } a \text{ in state } s \text{ is \emph{unsafe} iff }$  $$val_{\varphi,\mathcal{M}}(s,a,h)<\lambda_{sh}\cdot optval_{\varphi,\mathcal{M}}(s,h).$$
We refer to actions that are not unsafe as safe actions.

The task of the shield is to block any unsafe action from the agent, thereby restricting the set of available actions $\Act(s)$
to the set of safe actions.
A shield is a relation $\shielded{\pi} : \states \rightarrow 2^{\Act(s)}$ allowing at least one action for any state.

\section{Learned Shields for Safe RL}
\label{sec:method}
In this section, we present our iterative approach for safe reinforcement learning via automata learning and shielding.
We first discuss the setting in which the RL agents operates
and give our problem statement.
Then we give an overview of our approach. Finally, we
discuss the individual steps of our approach
in detail. 

\subsection{Setting and Problem Statement}

\textbf{Setting.} We consider an RL agent acting in an unknown environment
that can be modeled as an MDP $\mathcal{M} = \langle \states,s_0,\Act, \trans\rangle$
with an associated reward function $\rewards : \states \times \Act \times \states \rightarrow \mathbb{R}$.
However, since the environment is unknown at the beginning of the learning phase, the agent has no knowledge about the structure of $\mathcal{M}$.

We assume that the safety critical properties are given in form of an
LTL formula $\varphi$. Without knowledge about the safety-relevant dynamics of the environment, it is not possible to prohibit violating $\varphi$ while the RL agent is exploring the environment.

During the exploration phase of the RL agent, a multiset of reward traces is collected. We assume to have an observation function $Z : \states \rightarrow O$ given that maps any state $s\in \states$ states to a safety-relevant observation $o\in O$. 

\noindent\textbf{Problem Statement.}
Consider the setting as discussed above. The goal of our approach is to use
the available data from the environment in form of collected traces
and the given safety specification
to prevent safety violations if possible.

\subsection{Overview of Iterative Safe RL via Learned Shields}
\begin{figure}[t]
    \centering
   \input{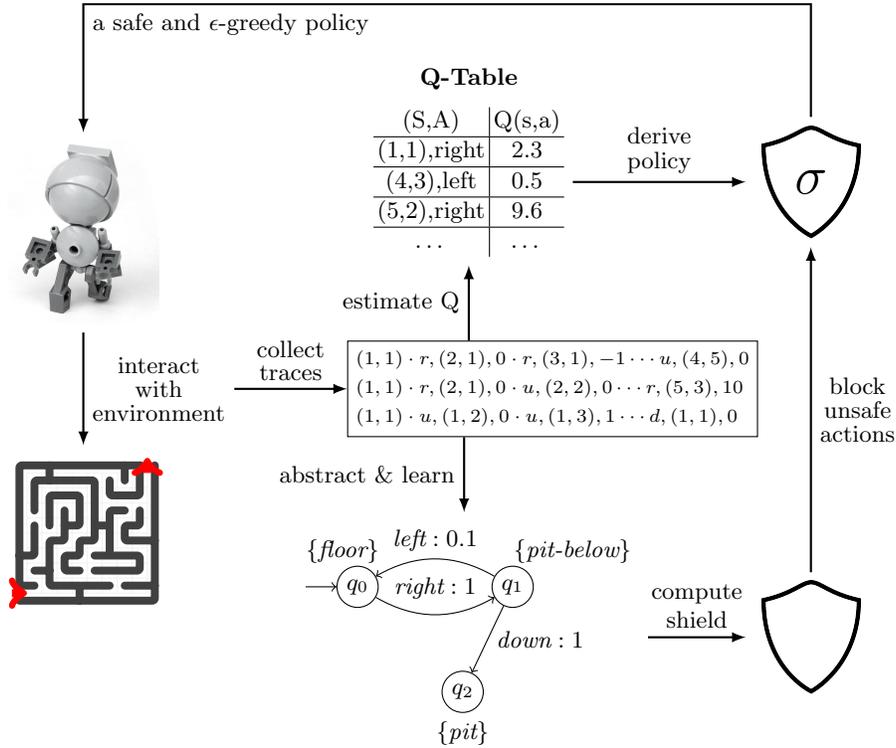}
   \vspace{-0.5cm}
    \caption{Iterative safe reinforcement learning via learned shields.}
    \label{fig:overview}
\end{figure}
Our approach combines automata learning, shielding, and reinforcement learning in an iterative manner. 
Our approach performs $n_{iter}$ iterations.
At the first iteration with $i=1$, we start from an empty multiset of reward traces $\mathcal{T}=\emptyset$,
an initial learned $\shielded[0]{\mdp}$ with a single state and a self-loop for any action, and an initial shield 
$\shielded[0]{\shield}$ that allows at every step any action.

At each iteration $i$ of $n$ iterations, our approach works as follows:
\begin{description}
\item [Step 1 - Exploration.] The RL agent explores the environment to learn the optimal policy. At each iteration $i$, the agent 
trains for a number of $n_{episodes}$.
The agent is augmented by a shield $\shielded[i-1]{\pi}$ that restricts its available actions.
The learned MDP $\shielded[i-1]{\mdp}$ is simulated in parallel
during the exploration.
Each training episode yields a reward trace $\tau$ that is added to
the multiset of all collected traces $\mathcal{T}$, i.e., $\mathcal{T}'=\{\tau\}\uplus\mathcal{T}$.

\item [Step 2 - MDP Learning.] 
After executing $n_{episodes}$ episodes, we learn a deterministic labeled MDP $\shielded[i]{\mdp}$ from $\mathcal{T}$.

\item [Step 3 - Shield Construction.] Using $\shielded[i]{\mdp}$, a given specification $\varphi$, and a finite horizon $h$,
we compute a shield $\shielded[i]{\pi}$ and continue in \textbf{Step 1}.

\end{description}

Figure~\ref{fig:overview} provides a graphical overview of the proposed approach. Based on this figure, we discuss the individual steps of our approach in detail. For each iteration 
$1 \leq i \leq n_{iter}$ our approach performs the following steps.

\textbf{Step 1 - Exploration.} The agent interacts with an unknown stochastic environment $\mdp = \langle S,s_0,\Act, \trans  \rangle$, depicted by a maze. The agent picks actions that are considered to be safe by the current shield
and receives observations of the state of the environment
and rewards. 
At iteration $i$, we have given the set of collected reward traces
$\mathcal{T}$, the learned MDP $\shielded[i-1]{\mdp}
= \langle \shielded{S},\shielded[0]{s},\shielded{\Act}, \shielded{\trans}, L\rangle$,
and the current shield $\shielded[i-1]{\pi}$.
Each episode resets the environment and  starts in a fixed initial state $s_0\in \states$ for $\mathcal{M}$
and $\shielded[0]{s}\in \shielded{\states}$ for $\shielded{\mdp}$.
At every step $t$, the agent observes $s_t\in \states$.
Based on the observed label $Z(s_t)$, the learned MDP $\shielded[i-1]{\mdp}$ moves to state $\shielded[t]{s}$.
We give in Section~\ref{sec:detail-training} the details on how to simulate $\shielded[i]{\mdp}$.

The shield determines the set of safe actions $\shielded{\Act} = \shielded[i-1]{\pi}(\shielded[t]{s})$ and sends it to the agent.
The agent selects a next action $a_{t+1}\in \shielded{\Act}(s)$. The environment executes $a_{t+1}$
and sends the agent a reward $r_{t+1}$ and state observation $s_{t+1}$.
Based on $(s_t, a_{t+1}, r_{t+1}, s_{t+1})$, the agent performs a policy update. Figure~\ref{fig:overview} represents the learned Q-function of the RL agent as a Q-table, from which we derive an $\epsilon$-greedy policy for training.
Please note that our approach is general and applicable to deep Q-learning as well as to tabular Q-learning. 

A training episode ends after a maximal number of $t_{max}$ steps.
It may end earlier in case of violating safety or performing the task that needs to be learned (e.g., reaching a certain goal state).
Each episode yields a reward trace 
$\tau = s_0 a_1 r_1 s_1  \cdots  s_{n-1} a_n r_n s_n$ that is added to the multiset $\mathcal{T}$ of all collected traces, i.e., 
$\mathcal{T}' = \{\tau\} \uplus \mathcal{T}$.

\textbf{Step 2 - MDP Learning.}
After $n_{episodes}$ training episodes, resulting in a new reward trace per episode, we learn a new model of the environment dynamics.
Given the multiset of reward traces $\mathcal{T}$ observed while exploring the environment, we abstract away all information in the traces that is not relevant to safety. 
For each reward trace $\tau = s_0 a_1 r_1 s_1  \cdots  s_{n-1} a_n r_n s_n  \in \mathcal{T}$,
we first discard the rewards to obtain a path $\rho = s_0 a_1 s_1  \cdots  s_{n-1} a_n s_n$ and second apply the abstraction function $Z$
to obtain an observation trace $\tau_o = Z(\rho) = Z(s_0) a_1 Z(s_1)  \cdots  Z(s_{n-1}) a_n Z(s_n)$.
For example, the states in the path $\rho$ may represent the exact coordinates
of the agent's positions and distance measurements. Relevant for safety might only be the distances. In such a case, the abstraction function $Z$ would abstract away the concrete position and only keep the distances.


To compute the  deterministic labeled MDP $\shielded[i]{\mathcal{M}}$,
we invoke \IOALERGIA by calling
$\IOALERGIA{}(\mathcal{T}_0,\epsilon_\mathrm{\ALERGIA{}})$, where 
$\epsilon_\mathrm{\ALERGIA{}}$ is a parameter specifying the significance level
of statistical tests performed by \IOALERGIA.

Additionally, we make $\shielded[i]{\mdp}$ \emph{action-complete}.
During the training phase, we propose to make $\shielded[i]{\mdp}$ with $i<n$ action-complete
by adding self-loop transitions to all state-action pairs $(\shielded{s},\shielded{a})$ 
where $\shielded{\trans}(\shielded{s},\shielded{a})$ is not defined. That is, we set for all such
state-action pairs $\shielded{\trans}(\shielded{s},\shielded{a}) = \{\shielded{s} \mapsto 1\}$.
The rationale behind this is that whenever an action's effect is unknown,
we assume that it leaves the safety of the corresponding state unchanged.

For the final shield $\shielded{\pi}$
that will be used permanently after training,
we propose a more conservative approach and to make $\shielded[i]{\mdp}$ with
$i=n$ action-complete by adding a special sink state $s_{\lnot\varphi}$
that violates $\varphi$ and adding
transitions to $s_{\lnot\varphi}$.
In the training phase, the resulting shield $\shielded{\pi}$
would not be suitable since it would prohibit
exploration.
For the exploration phase, however, the behavior of $\shielded{\pi}$
may be desirable since it blocks
behavior that has not been explored sufficiently.

\textbf{Step 3 - Shield Construction.}
The abstract MDP $\shielded[i]{\mathcal{M}}$ encodes the safety-relevant information about the environment.
We use $\shielded[i]{\mathcal{M}}$, the safety specification $\varphi$,
and a finite horizon $h$, to compute the safety values of
all state-action pairs in $\shielded[i]{\mathcal{M}}$.
Based on a given relative threshold $\lambda_{sh}$,
we compute a shield $\shielded[i]{\pi}$ which allows 
for any state all actions that are safe w.r.t. $\lambda_{sh}$
and the optimal safety value. 
After constructing the shield $\shielded[i]{\pi}$, the current shield of the agent is set to 
$\shielded[i]{\pi}$. The agent continues to explore the environment and learn the optimal strategy using $\shielded[i]{\pi}$
(Step 1).

\subsection{Details for Training using Learned Shields} 
\label{sec:detail-training}
\begin{algorithm}[t]
\begin{algorithmic}[1]
\Require Q-function $Q$, a learned deterministic labeled MDP $\shielded{\mdp}
= \langle \shielded{S},\shielded{\Act},\shielded[0]{s}, \shielded{\trans}, L\rangle$, exploration rate $\epsilon$,
safety shield $\shielded{\pi}$, environment with $\step$ and $\reset$
\Ensure Updated Q-function $Q$
\State $\shielded{s} \gets \shielded[0]{s}$ \label{algline:init1}
\State $s \gets\reset$  \label{algline:init2}
\State $\mathit{t} \gets 0$  \Comment{steps}
\While{$\mathit{t} < \maxeplen$}
\State $\shielded{\Act} \gets \shielded{\pi}(\shielded{s})$ \Comment{safe actions for s} \label{algline:det_safe}
\If{$\coinFlip(\epsilon)$}
\State $a \gets \randSel(\shielded{\Act})$ \label{algline:random_exp}
\Else
\State $a \gets \argmax_{a' \in \shielded{\Act}}Q(s,a')$ \label{algline:exploit}
\EndIf
\State $r,s' \gets \step(a)$ \Comment{step in environment} \label{algline:step}
\State $\shielded{s} \gets \shielded{s'} \text{ where } 
 \trans(\shielded{s},a)(\shielded{s'}) > 0 \text{ and } L(\shielded{s'}) = Z(s')$
 \Comment{step in learned MDP} \label{algline:step_aut}
 \State $\textsc{UpdateQ}(s, a, r, s')$
 \If{$s'$ is a terminal state}\label{algline:goal}
 \State \textbf{break}
 \EndIf
 \State $t \gets t + 1$
\EndWhile
\end{algorithmic}
\caption{A single RL training's episode using a learned shield.}
\label{alg:rl_episode}
\end{algorithm}

In this section we discuss a single RL episode of 
an agent augmented with a learned shield in detail.
The pseudocode is given in Algorithm~\ref{alg:rl_episode}.
The RL agent interacts with the environment through two operations: $\reset$ and $\step$\footnote{the convention of OpenAI gym~\cite{DBLP:journals/corr/BrockmanCPSSTZ16}}.
\begin{description}
\item [$\reset$:] This operation resets the environment to a fixed initial state.
This state $s_0$ from the unknown environment MDP is also returned from $\reset$.
\item [$\step$:] The operation $\step(a)$
takes an action $a$ and executes $a$ causing a probabilistic 
state transition. It returns a pair $r,s'$, where $r$ is the reward gained by performing $a$ and $s'$ is the reached state when executing $a$.
\end{description} 

The algorithm starts with initializing
the learned MDP state as well as the environment state (Line~\ref{algline:init1}
and \ref{algline:init2}). 
Then we enter a loop in which the agent performs a maximum of 
$\maxeplen$ steps.

The RL agent applies $\epsilon$-greedy 
learning, i.e., it explores a random action with probability $\epsilon$
and otherwise performs the optimal action according to the
RL agent's current knowledge. Note that both, random exploration and exploitation,
are shielded, i.e., actions are chosen from the set of safe actions. 

For every step, the shield provides a set of safe actions
(Line~\ref{algline:det_safe}).
With probability $\epsilon$, the RL agent selects a random safe
action in Line~\ref{algline:random_exp}. Otherwise, it determines the currently
optimal action in Line~\ref{algline:exploit}.
The chosen action is executed in Line~\ref{algline:step}.
In Line~\ref{algline:step_aut}, the state  
of $\shielded{\mdp}$ is updated. To conclude the training step, we update the agent's $Q$-table.

If the agents visits a terminal state, the loop terminates before performing $\maxeplen$ steps (Line ~\ref{algline:goal}). A terminal state may, for instance, be
reached by completing the task to be solved or by violating
safety.

\textbf{Execution phase.}
After training, we use the same approach to execute an agent, but with $\epsilon$ set to $0$,
such that only safe and optimal actions are executed. 


\section{Experiments}
\label{sec:experiments}
\label{sec:eval}

In our experimental evaluation, we evaluated our approach on
$24$ slippery gridworld environments of varying shapes and sizes.
We implemented a tabular Q-learning agent that should learn to reach a given goal state quickly while staying safe.
We learn deterministic labeled MDPs using \IOALERGIA implemented in \AALpy~\cite{aalpy}. The shields are created from the learned MDPs
using the shield synthesis tool \tempest~\cite{tempest}.
In our experiments, we discuss the scalability of our approach and
its effectiveness by comparing the averaged gained reward during training with and without learned shields.

\noindent \textbf{Experimental setup.}
All experiments have been executed on a
desktop computer with a 4 x 2.70 GHz Intel Core i7-7500U CPU, 7.7 GB of RAM running Arch Linux.

\noindent \textbf{Availability.}
An implementation of our framework for iterative safe RL via learned shields is available online,
together with several examples and detailed execution instructions to reproduce our results\footnote{\url{https://github.com/DES-Lab/Automata-Learning-meets-Shielding}}.
The implementation includes a shielded tabular Q-learning agent written in Python.

\noindent \textbf{Usability.}
Our prototype implementation allows users to easily perform their own experiments. \AALpy's \IOALERGIA implementation has a simple text-based interface and \tempest uses the well-established PRISM MDP format that is also supported by \AALpy, as well as a property language similar to PRISM and Storm. 
Furthermore, our prototype implementation can be easily extended. The tools, \AALpy and \tempest, can be used as black boxes. Therefore, there is no need to know the implementation details of these tools.
The only caveat is that \tempest is easiest to use through a Docker container.

\subsection{Case Study Subjects}

\noindent\textbf{Gridworlds.}
We used three types of parameterized gridworlds, the smallest instances are depicted in
Figure~\ref{fig:zigzaggw}, Figure~\ref{fig:slipperygw} and Figure~\ref{fig:wallgw}.
Each gridworld has a dedicated \emph{start} tile and a \emph{goal} tile, marked by \emph{Entry} and \emph{Goal} on
the left-hand side of the figures.
A gridworld might also have \emph{intermediate-goal} tiles.
If a tile is marked as \emph{pit}, then this tile marks an unsafe location and the agent is not allowed to visit this tile.
Black tiles represent \emph{walls} that restrict movement but are not
safety critical.
Additionally, each tile has a \emph{terrain}, denoted by
lowercase letters on the right-hand side of the figures.
If a tile is ``slippery'', the tile is labeled with an arrow and grey gradients on the left-hand side of the figures.
If an agent tries to move from a slippery tile,
the intended movement of the agent might be altered
into the direction of the arrow with a specific tile-dependent probability. The length of arrows corresponds to the probability of slipping. That is, a long arrow pointing downward left means that the agent is very likely to slip either to the left or downwards.
Additionally, every tile has $(x,y)$ coordinates.

\def\SCALING{0.6}
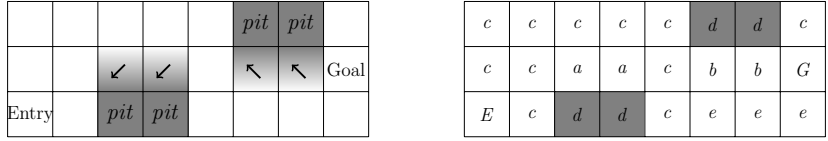
\begin{figure}[t!]
  \centering
  \begin{minipage}{.5\textwidth}
    \centering
    \scalebox{\SCALING}{
      \begin{tikzpicture}
        [box/.style={rectangle,draw=black,thick}]

        \pit{2}{0} \pit{3}{0}
        \pit{5}{2} \pit{6}{2}
        \slipdown{2}{1}
        \slipdown{3}{1}
        \slipup{5}{1}
        \slipup{6}{1}
        \draw (0.5,0.5) node{\large{Entry}};
        \draw (7.5,1.5) node{\large{Goal}};
        \draw[step=1cm,black,thin] (0,0) grid (8,3);
      \end{tikzpicture}
    }

  \end{minipage}%
  \begin{minipage}{0.5\textwidth}
    \centering
    \scalebox{\SCALING}{
      \begin{tikzpicture}
        \celllab{0}{0}{white}{E}
        \celllab{0}{1}{white}{c}
        \celllab{0}{2}{white}{c}

        \celllab{1}{0}{white}{c}
        \celllab{1}{1}{white}{c}
        \celllab{1}{2}{white}{c}

        \celllab{2}{0}{gray}{d}
        \celllab{2}{1}{white}{a}
        \celllab{2}{2}{white}{c}

        \celllab{3}{0}{gray}{d}
        \celllab{3}{1}{white}{a}
        \celllab{3}{2}{white}{c}

        \celllab{4}{0}{white}{c}
        \celllab{4}{1}{white}{c}
        \celllab{4}{2}{white}{c}

        \celllab{5}{0}{white}{e}
        \celllab{5}{1}{white}{b}
        \celllab{5}{2}{gray}{d}

        \celllab{6}{0}{white}{e}
        \celllab{6}{1}{white}{b}
        \celllab{6}{2}{gray}{d}

        \celllab{7}{0}{white}{e}
        \celllab{7}{1}{white}{G}
        \celllab{7}{2}{white}{c}

        \draw[step=1cm,black,thin] (0,0) grid (8,3);
      \end{tikzpicture}
    }
  \end{minipage}
  \caption{The smallest of the zigzag gridworlds.}
  \label{fig:zigzaggw}
\end{figure}
\def\SCALING{0.6}

\begin{figure}[t!]
  \centering
  \begin{minipage}{.5\textwidth}
    \centering
    \scalebox{\SCALING}{
      \begin{tikzpicture}
        [box/.style={rectangle,draw=black,thick}]

        \pit{2}{0} \pit{3}{0}

        \slipdownhard{1}{0}
        \slipdownhard{2}{1}
        \slipdownhard{3}{1}
        \slipdownhard{4}{1}
        \slipdownhard{4}{0}
        \slipdown{1}{1}
        \slipdown{5}{0}
        \slipdown{5}{1}
        \slipdown{1}{2}
        \slipdown{2}{2}
        \slipdown{1}{3}
        \slipdown{2}{3}

        \pit{5}{5}
        \pit{6}{5}
        \slipuphard{7}{5}
        \slipuphard{4}{5}
        \slipuphard{7}{4}
        \slipuphard{6}{4}
        \slipuphard{5}{4}
        \slipuphard{4}{4}
        \slipup{7}{3}
        \slipup{6}{3}
        \slipup{5}{3}

        \slipleft{2}{4}
        \slipleft{3}{5}
        \slipleft{3}{4}
        \slipleft{3}{3}
        \slipleft{3}{2}
        \slipleft{4}{3}
        \slipleft{4}{2}
        \slipleft{5}{2}
        \slipleft{6}{2}
        \slipleft{6}{1}
        \slipleft{7}{2}
        \slipleft{7}{1}
        \slipleft{8}{1}

        \draw (0.5,0.5) node{\large{Entry}};
        \draw (8.5,4.5) node{\large{Goal}};

        \draw[step=1cm,black,thin] (0,0) grid (9,6);
      \end{tikzpicture}
    }

  \end{minipage}%
  \begin{minipage}{0.5\textwidth}
    \centering
    \scalebox{\SCALING}{
      \begin{tikzpicture}
        \celllab{0}{5}{white}{g}
        \celllab{0}{4}{white}{g}
        \celllab{0}{3}{white}{g}
        \celllab{0}{2}{white}{g}
        \celllab{0}{1}{white}{g}
        \celllab{0}{0}{white}{E}

        \celllab{1}{5}{white}{g}
        \celllab{1}{4}{white}{g}
        \celllab{1}{3}{white}{b}
        \celllab{1}{2}{white}{b}
        \celllab{1}{1}{white}{b}
        \celllab{1}{0}{white}{a}

        \celllab{2}{5}{white}{g}
        \celllab{2}{4}{white}{c}
        \celllab{2}{3}{white}{c}
        \celllab{2}{2}{white}{c}
        \celllab{2}{1}{white}{a}
        \celllab{2}{0}{gray}{d}

        \celllab{3}{5}{white}{c}
        \celllab{3}{4}{white}{c}
        \celllab{3}{3}{white}{c}
        \celllab{3}{2}{white}{c}
        \celllab{3}{1}{white}{a}
        \celllab{3}{0}{gray}{d}

        \celllab{4}{5}{white}{e}
        \celllab{4}{4}{white}{e}
        \celllab{4}{3}{white}{c}
        \celllab{4}{2}{white}{c}
        \celllab{4}{1}{white}{a}
        \celllab{4}{0}{white}{a}

        \celllab{5}{5}{gray}{d}
        \celllab{5}{4}{white}{e}
        \celllab{5}{3}{white}{f}
        \celllab{5}{2}{white}{c}
        \celllab{5}{1}{white}{b}
        \celllab{5}{0}{white}{b}

        \celllab{6}{5}{gray}{d}
        \celllab{6}{4}{white}{e}
        \celllab{6}{3}{white}{f}
        \celllab{6}{2}{white}{c}
        \celllab{6}{1}{white}{c}
        \celllab{6}{0}{white}{g}

        \celllab{7}{5}{white}{e}
        \celllab{7}{4}{white}{e}
        \celllab{7}{3}{white}{f}
        \celllab{7}{2}{white}{c}
        \celllab{7}{1}{white}{c}
        \celllab{7}{0}{white}{g}

        \celllab{8}{5}{white}{g}
        \celllab{8}{4}{white}{G}
        \celllab{8}{3}{white}{g}
        \celllab{8}{2}{white}{g}
        \celllab{8}{1}{white}{c}
        \celllab{8}{0}{white}{g}

        \draw[step=1cm,black,thin] (0,0) grid (9,6);
      \end{tikzpicture}
    }
  \end{minipage}
  \caption{The smallest of the slippery shortcut gridworlds.}
  \label{fig:slipperygw}

\end{figure}
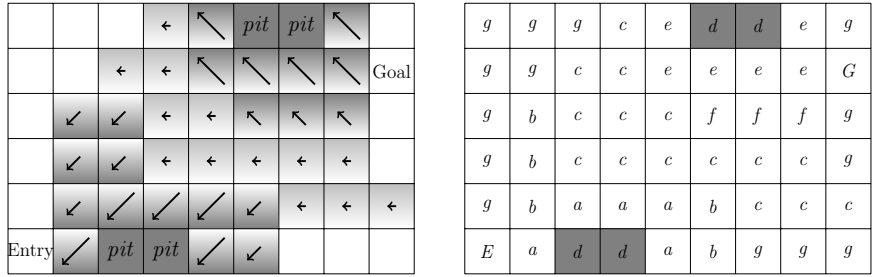
\def\SCALING{0.6}

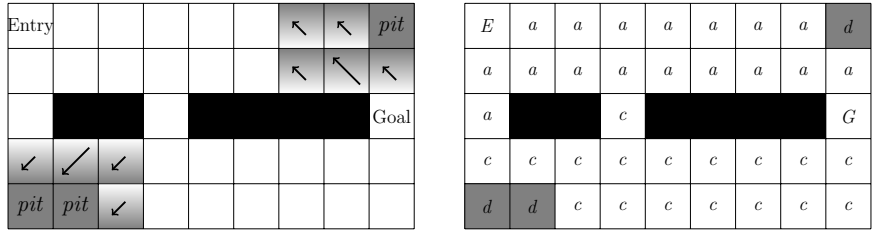
\begin{figure}[t!]
  \centering
  \begin{minipage}{.5\textwidth}
    \centering
    \scalebox{\SCALING}{
      \begin{tikzpicture}
        [box/.style={rectangle,draw=black,thick}]

        \pit{0}{0} \pit{1}{0}
        \slipdown{0}{1}
        \slipdown{2}{1}
        \slipdown{2}{0}
        \slipdownhard{1}{1}

        \pit{8}{4}
        \slipup{8}{3}
        \slipup{6}{3}
        \slipup{6}{4}
        \slipup{7}{4}
        \slipuphard{7}{3}
        \draw (0.5,4.5) node{\large{Entry}};
        \draw (8.5,2.5) node{\large{Goal}};

        \wall{1}{2}
        \wall{2}{2}

        \wall{4}{2}
        \wall{5}{2}
        \wall{6}{2}
        \wall{7}{2}

        \draw[step=1cm,black,thin] (0,0) grid (9,5);
      \end{tikzpicture}
    }

  \end{minipage}%
  \begin{minipage}{0.5\textwidth}
    \centering
    \scalebox{\SCALING}{
      \begin{tikzpicture}
        \celllab{0}{4}{white}{E}
        \celllab{0}{3}{white}{a}
        \celllab{0}{2}{white}{a}
        \celllab{0}{1}{white}{c}
        \celllab{0}{0}{gray}{d}

        \celllab{1}{4}{white}{a}
        \celllab{1}{3}{white}{a}
        \celllab{1}{2}{white}{c}
        \celllab{1}{1}{white}{c}
        \celllab{1}{0}{gray}{d}

        \celllab{2}{4}{white}{a}
        \celllab{2}{3}{white}{a}
        \celllab{2}{2}{white}{c}
        \celllab{2}{1}{white}{c}
        \celllab{2}{0}{white}{c}

        \celllab{3}{4}{white}{a}
        \celllab{3}{3}{white}{a}
        \celllab{3}{2}{white}{c}
        \celllab{3}{1}{white}{c}
        \celllab{3}{0}{white}{c}

        \celllab{4}{4}{white}{a}
        \celllab{4}{3}{white}{a}
        \celllab{4}{2}{white}{c}
        \celllab{4}{1}{white}{c}
        \celllab{4}{0}{white}{c}

        \celllab{5}{4}{white}{a}
        \celllab{5}{3}{white}{a}
        \celllab{5}{2}{white}{c}
        \celllab{5}{1}{white}{c}
        \celllab{5}{0}{white}{c}

        \celllab{6}{4}{white}{a}
        \celllab{6}{3}{white}{a}
        \celllab{6}{2}{white}{c}
        \celllab{6}{1}{white}{c}
        \celllab{6}{0}{white}{c}

        \celllab{7}{4}{white}{a}
        \celllab{7}{3}{white}{a}
        \celllab{7}{2}{white}{c}
        \celllab{7}{1}{white}{c}
        \celllab{7}{0}{white}{c}

        \celllab{8}{4}{gray}{d}
        \celllab{8}{3}{white}{a}
        \celllab{8}{2}{white}{G}
        \celllab{8}{1}{white}{c}
        \celllab{8}{0}{white}{c}

        \wall{1}{2}
        \wall{2}{2}

        \wall{4}{2}
        \wall{5}{2}
        \wall{6}{2}
        \wall{7}{2}

        \draw[step=1cm,black,thin] (0,0) grid (9,5);
      \end{tikzpicture}
    }
  \end{minipage}
  \caption{The smallest of the wall gridworlds.}
  \label{fig:wallgw}
\end{figure}

\paragraph{Gridworld Shapes.}
Next, we discuss each of the three gridworld shapes briefly,
where Figures~\ref{fig:zigzaggw} to \ref{fig:wallgw} show the smallest
gridworld of each shape.
To get insights into how the state space affects learning, we vary each type of gridworld in size by creating eight versions of increasing size.
\begin{itemize}
    \item {\em zigzag}: In the \emph{zigzag} gridworlds illustrated
    in Figure~\ref{fig:zigzaggw}, we have pairs
    of pit tiles located in alternation at the bottom and the top of the map
    with a fixed distance between adjacent pit pairs. The goal is placed so that
    the agent could move along a straight line from left to right, but it would travel across
    slippery tiles next to the pits. The shield thus helps to avoid these
    dangerous tiles to perform zigzag walks to the goal.
    \item {\em slippery shortcuts}: The \emph{slippery shortcuts} gridworlds
    illustrated in Figure~\ref{fig:slipperygw} are similar to the \emph{zigzag}
    maps in that the agent could take shortcuts across slippery tiles
    next to pits. In this case, the probability of unsuccessful moves
    decreases with increasing distance from the pits.
    Hence, an optimal policy has to find a balance
    between taking risks and gaining higher reward due to shorter paths.
    \item {\em walls}: In the \emph{walls} gridworlds illustrated
    in Figure~\ref{fig:wallgw}, the agents must find a
    way from the start to the goal by navigating around walls and pits
    that block the shortest paths. There are fewer slippery tiles.
\end{itemize}


\noindent
\textbf{Reinforcement Learning.}
We implemented a tabular Q-learning agent.
The agent's task is to navigate from start to goal by moving into one of the four cardinal directions at each time step.

Every training episode, the agent collects reward traces of the form $\tau = (x_0, y_0), a_1, r_1, (x_1, y_1)\dots
a_n, r_n, (x_n, y_n)$ with $x_i$ and $y_i$ representing the $x$ and $y$ coordinates of the tiles.
The set of available \emph{actions}
comprises of the four actions $\{$left, right, up, down$\}$ to move into the corresponding direction.
The \emph{reward function} of the agent is defined as follows:
\begin{itemize}
    \item If the agent reaches the goal within less than $t_{max}$ steps, it receives a reward of $+100$. Additionally, reaching the goal ends the training episode.
    \item If the agent reaches a tile with a \emph{pit}, it receives a punishment of $-100$ and the training episode ends.
    \item Additionally, the agent receives a reward of $-0.5$ per step.
    \item Gridworlds might have intermediate goal states that are rewarded with $+20$ but do not terminate the episode.
\end{itemize}

We use the following \emph{learning parameters} for all experiments:
The tabular Q-learning agent was trained with a learning rate of $\alpha = 0.1$, a discount factor of $\gamma = 0.9$, and an initial exploration rate of $\epsilon = 0.4$ throughout all experiments. We chose an exponential epsilon decay of $\epsilon' = 0.9999 \cdot \epsilon$ with an update after every learning episode.

\textbf{MDP Learning.}
To get the observations for MDP learning, we perform
an abstraction over the states via the function $Z$. Given a concrete state $(x,y)$, the function
$Z((x,y))$ maps to a pair $(\mathit{terr},\mathit{Pit})$, where $\mathit{terr}$ is the terrain of the tile at $(x,y)$
and $\mathit{Pit}$ is a set of propositions denoting whether a pit is located in the neighboring tiles in each of the four cardinal directions.

For example, if the agent is at the coordinate $(1,0)$
in Figure~\ref{fig:zigzaggw}, with the origin of coordinates on the
bottom left, the abstract observation would be $Z((1,0)) =
(c, \{\textit{pit-right}\})$. The terrain is $c$ and there
is a pit on the right.

\textbf{Shield Construction.}
In all experiments, visiting a pit represents a safety violation.
This property can be represented in LTL as follows: $\varphi = \textbf{G}(\lnot \mathit{pit})$.
Using this specification $\varphi$ and a deterministic labeled MDP
$\shielded{\mdp}$, we compute a shield $\shielded{\pi}$ using
a relative threshold $\lambda_{sh}=0.95$ and a finite horizon of $h=2$.
Thus, for any given state, the resulting shield $\shielded{\pi}$
allows an action $a$ if the probability of not falling into a pit within the next 2 steps is at least $0.95 \cdot \alpha$, with $\alpha$ being the probability of not falling into a pit when taking the optimal action $a'$.

\subsection{Experimental Results}
In the following, we report on the performance of RL agents augmented with learned shields compared to the performance of unshielded RL agents.

To account for the stochastic
nature of the environment and RL, we repeat every experiment $30$ times.
For each experiment, the number of iterations is set to $n_{iter}=30$,
the number of training episodes per iteration is $n_{episodes}=1000$, and the maximal length of a training episode is set to $t_{max}=200$ steps.

To evaluate the learned policies, we execute the
policies at various stages throughout training and compare their
performance.
For every iteration, i.e., after every $1000$\textsuperscript{th} training episode, we evaluate the intermediate policies of
the agents by setting the exploration
rate $\epsilon$ to $0$ and performing $1000$ episodes.
Over these $1000$ episodes, we compute the average cumulative reward
of the intermediate policy. We refer to this value as \emph{return}.
For shielded agents, their corresponding shields are used
during the intermediate executions for evaluation.

\begin{figure}[t]
    \centering
    \input{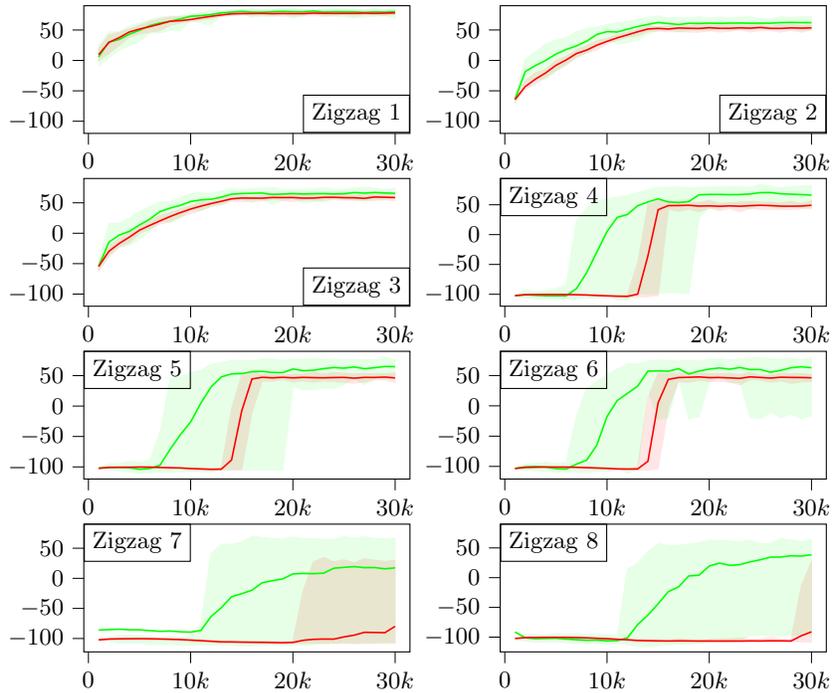}
    \caption{The return gained by intermediate policies throughout
    reinforcement learning in the \emph{zigzag} gridworlds. The x-axes
    display the return and the y-axes display the episodes
    at which policies are evaluated.
    The green plots represent shielded performance, whereas
    the red plots represent unshielded performance.}
    \label{fig:gridnworld_1_results}
\end{figure}
\begin{figure}[h]
    \centering
    \input{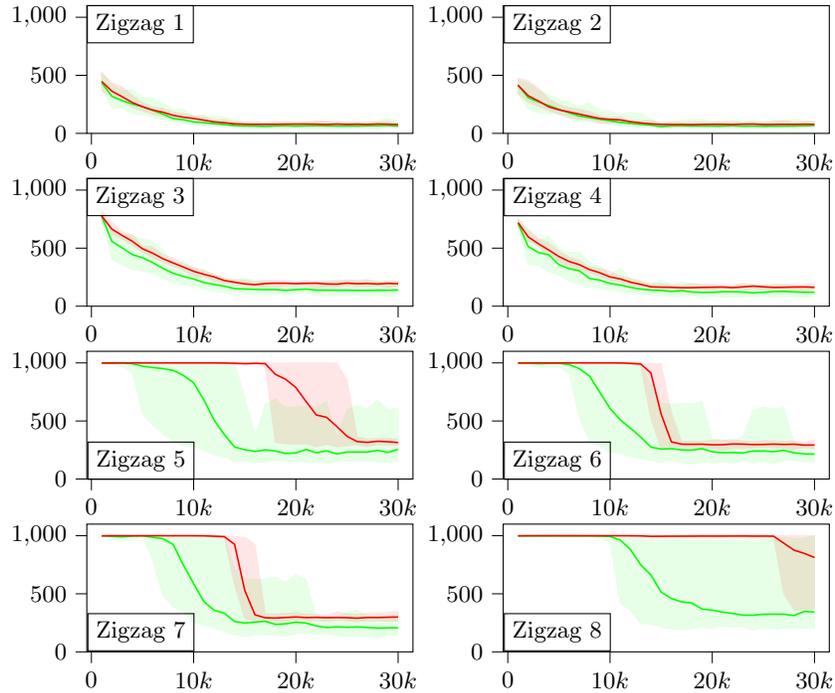}
    \caption{The number of safety violations throughout RL in the \emph{zigzag} gridworlds. The x-axes display the number of violations and the y-axes display the episodes at which the policies are evaluated. The green plots represent the number of violations under the shielded policies, whereas
    the red plots represent the same under unshielded policies.}
    \label{fig:pitfrequency_1_results}
\end{figure}

Figure~\ref{fig:gridnworld_1_results}, Figure~\ref{fig:gridnworld_2_results}, and Figure~\ref{fig:gridnworld_3_results} show
plots of the results of this evaluation, where the x-axes display the episodes
and the y-axes display the return. The average performance of the \emph{shielded
agents} is represented by a thick \emph{green line}, whereas a thick \emph{red} line represents
the average performance of \emph{unshielded agents}. The light green and light red areas depict
the range between the minimum and maximum performance of the shielded and the unshielded
agents, respectively. 
Figure~\ref{fig:pitfrequency_1_results}, Figure~\ref{fig:pitfrequency_2_results}, and Figure~\ref{fig:pitfrequency_3_results} depict the number of safety violations throughout training, where the x-axes display the episodes and the y-axes display the number of times the agent visited a pit.
The average number of safety violations of the \emph{shielded
agents} are represented by a thick \emph{green line}, whereas a thick \emph{red} line represents
the average number of safety violations of \emph{unshielded agents}. The light green and light red areas depict
the range between the minimum and maximum number of safety violations of the shielded and the unshielded
agents, respectively. 
In the following, we discuss the
results from the experiments with the different gridworld shapes.

\subsection{Zigzag Gridworlds}
We start by discussing the performance of the RL agents in
the \emph{zigzag} gridworlds; see Figure~\ref{fig:zigzaggw} for the smallest such
environment. Figure~\ref{fig:gridnworld_1_results} shows the
return, and Figure~\ref{fig:pitfrequency_1_results} shows the number of safety violations at various stages
of RL.

The experiments in Figure~\ref{fig:gridnworld_1_results} show almost identical returns and number of safety violations for shielded and unshielded agents for the three smallest gridworld instances.
Starting with the fourth-smallest
gridworld, shielded RL performs better on average.
Initially, during the first episodes of both RL configurations, the average
return is approximately $-100$, which is the penalty
for falling into a pit. This means that the agent
consistently falls into pits in the early stages of learning.
After approximately $7$ iterations, i.e, at episode $7000$,
the shielded agents start to reach the goal states, which leads to an increase in the return and a decrease in the number of safety violations.
Unshielded agents need about twice the time to reach the goal location.
Similar observations can be made for the next larger
environments \emph{Zigzag 5} and \emph{Zigzag 6}, too.
For the two largest \emph{zigzag} gridworlds, unshielded RL fails to consistently reach the goal after $30,000$ training episodes. These two environments require relatively long paths
to be traversed and the gained rewards are sparse.
Hence, learned safety shields may benefit RL in environments with sparse rewards, where safety violations may prevent the agents
from visiting states that give a positive reward.
The decreases in the number of safety violations, as shown in Figure~\ref{fig:pitfrequency_1_results}, match the observations on the performance increases illustrated in Figure~\ref{fig:gridnworld_1_results}.

On the negative side, note that the growth of the return is steeper for unshielded RL. For example, considering the environment \emph{Zigzag 4}, it takes about $10000$ episodes to reach an average return greater than $0$ in the shielded case and it takes $15000$ episodes
in the unshielded case. Hence, there are $3000$ episodes between
first reaching the goal and reaching it more consistently for shielded
RL, whereas unshielded RL only requires $1000$ episodes to make
this jump in performance.

Furthermore, consider the range between the minimum and the maximum return that is depicted by the shaded areas in the figures.
The minimum return of unshielded RL is often lower than the
minimum return of shielded RL even though it performs better on average. There is more variance in the return obtained
by shielded RL.
Also, the range between the minimum and the maximum number of safety violations is high for \emph{Zigzag 8} until the end of training.
This could
result from learning of MDPs that do not sufficiently capture
safety-relevant information. We leave a closer investigation to future work.

\begin{figure}[t]
    \centering
\begin{tikzpicture}

\definecolor{darkgray176}{RGB}{176,176,176}

\pgfplotsset{every axis title/.append style={below right,at={(0,0.88), font=\small},draw=black}}

\begin{groupplot}[group style={group size=2 by 1,horizontal sep=1.2cm,vertical sep=0.6cm}]
\nextgroupplot[
tick align=outside,
tick pos=left,
title={Slippery 1},
x grid style={darkgray176},
xmin=-450, xmax=31450,
xtick style={color=black},
y grid style={darkgray176},
 xtick={0,10000,20000,30000},  xticklabels={$0$,$10k$,$20k$,$30k$},  scaled x ticks=false,      height=0.17\textheight, width=0.485\textwidth,
ymin=-100, ymax=125,
ytick style={color=black}
]
\path [draw=green, fill=green, opacity=0.1]
(axis cs:1000,-17.0065065065065)
--(axis cs:1000,-23.7808)
--(axis cs:2000,6.5125)
--(axis cs:3000,20.99)
--(axis cs:4000,31.479)
--(axis cs:5000,34.085)
--(axis cs:6000,46.4135)
--(axis cs:7000,49.603)
--(axis cs:8000,54.1775)
--(axis cs:9000,58.2665)
--(axis cs:10000,61.5255)
--(axis cs:11000,67.3325)
--(axis cs:12000,67.943)
--(axis cs:13000,71.0555)
--(axis cs:14000,72.2785)
--(axis cs:15000,75.451)
--(axis cs:16000,73.523)
--(axis cs:17000,70.9785)
--(axis cs:18000,71.768)
--(axis cs:19000,71.8515)
--(axis cs:20000,72.474)
--(axis cs:21000,71.3325)
--(axis cs:22000,71.434)
--(axis cs:23000,69.4465)
--(axis cs:24000,72.734)
--(axis cs:25000,73.8555)
--(axis cs:26000,75.0575)
--(axis cs:27000,73.706)
--(axis cs:28000,71.828)
--(axis cs:29000,75.0815)
--(axis cs:30000,71.568)
--(axis cs:30000,77.0415)
--(axis cs:30000,77.0415)
--(axis cs:29000,75.686)
--(axis cs:28000,75.055)
--(axis cs:27000,77.0205)
--(axis cs:26000,77.804)
--(axis cs:25000,78.846)
--(axis cs:24000,75.8375)
--(axis cs:23000,77.6725)
--(axis cs:22000,75.0315)
--(axis cs:21000,74.235)
--(axis cs:20000,75.285)
--(axis cs:19000,73.386)
--(axis cs:18000,75.5015)
--(axis cs:17000,75.6895)
--(axis cs:16000,77.894)
--(axis cs:15000,76.1685)
--(axis cs:14000,75.5805)
--(axis cs:13000,74.357)
--(axis cs:12000,68.7285)
--(axis cs:11000,69.793)
--(axis cs:10000,64.4725)
--(axis cs:9000,64.2205)
--(axis cs:8000,56.1905)
--(axis cs:7000,51.0455)
--(axis cs:6000,52.7625)
--(axis cs:5000,42.5235)
--(axis cs:4000,38.629)
--(axis cs:3000,23.1225)
--(axis cs:2000,8.788)
--(axis cs:1000,-17.0065065065065)
--cycle;

\path [draw=red, fill=red, opacity=0.1]
(axis cs:1000,-20.8173173173173)
--(axis cs:1000,-39.2422)
--(axis cs:2000,10.289)
--(axis cs:3000,19.1925)
--(axis cs:4000,31.3465)
--(axis cs:5000,40.3555)
--(axis cs:6000,49.2795)
--(axis cs:7000,51.89)
--(axis cs:8000,57.119)
--(axis cs:9000,57.747)
--(axis cs:10000,64.1345)
--(axis cs:11000,69.8845)
--(axis cs:12000,71.867)
--(axis cs:13000,71.0125)
--(axis cs:14000,72.3095)
--(axis cs:15000,72.3725)
--(axis cs:16000,76.022)
--(axis cs:17000,74.6405)
--(axis cs:18000,74.4285)
--(axis cs:19000,75.502)
--(axis cs:20000,73.1035)
--(axis cs:21000,73.692)
--(axis cs:22000,75.007)
--(axis cs:23000,75.6135)
--(axis cs:24000,73.882)
--(axis cs:25000,73.6945)
--(axis cs:26000,74.481)
--(axis cs:27000,75.194)
--(axis cs:28000,71.4095)
--(axis cs:29000,74.612)
--(axis cs:30000,75.42)
--(axis cs:30000,78.3235)
--(axis cs:30000,78.3235)
--(axis cs:29000,78.2275)
--(axis cs:28000,80.82)
--(axis cs:27000,76.15)
--(axis cs:26000,77.2525)
--(axis cs:25000,75.122)
--(axis cs:24000,77.438)
--(axis cs:23000,77.7105)
--(axis cs:22000,78.7745)
--(axis cs:21000,80.13)
--(axis cs:20000,75.8555)
--(axis cs:19000,77.546)
--(axis cs:18000,78.26)
--(axis cs:17000,75.4165)
--(axis cs:16000,80.195)
--(axis cs:15000,79.387)
--(axis cs:14000,77.173)
--(axis cs:13000,77.455)
--(axis cs:12000,72.3445)
--(axis cs:11000,71.2805)
--(axis cs:10000,66.6365)
--(axis cs:9000,64.626)
--(axis cs:8000,61.2525)
--(axis cs:7000,56.7365)
--(axis cs:6000,53.0215)
--(axis cs:5000,49.674)
--(axis cs:4000,36.861)
--(axis cs:3000,28.154)
--(axis cs:2000,12.761)
--(axis cs:1000,-20.8173173173173)
--cycle;

\addplot [semithick, green]
table {%
1000 -20.297
2000 7.50033
3000 22.25
4000 35.2725
5000 38.6547
6000 49.5503
7000 50.4213
8000 54.9363
9000 61.4867
10000 63.3338
11000 68.3505
12000 68.2283
13000 72.7797
14000 73.8903
15000 75.8232
16000 75.4365
17000 72.6258
18000 73.7468
19000 72.4357
20000 73.5013
21000 72.5958
22000 73.7257
23000 72.9472
24000 73.7908
25000 75.8405
26000 76.2567
27000 75.261
28000 73.3997
29000 75.4223
30000 74.574
};
\addplot [semithick, red]
table {%
1000 -29.2561
2000 11.535
3000 23.9792
4000 33.6595
5000 43.775
6000 51.2837
7000 54.8218
8000 59.7663
9000 61.0975
10000 65.4242
11000 70.4322
12000 72.1283
13000 73.2977
14000 75.0255
15000 75.3805
16000 78.1135
17000 75.1333
18000 76.4007
19000 76.4367
20000 74.4815
21000 76.6782
22000 77.437
23000 76.4478
24000 75.8013
25000 74.448
26000 75.5895
27000 75.6917
28000 76.3932
29000 76.472
30000 76.554
};

\nextgroupplot[
tick align=outside,
tick pos=left,
title={Slippery 8},
x grid style={darkgray176},
xmin=-450, xmax=31450,
xtick style={color=black},
y grid style={darkgray176},
 xtick={0,10000,20000,30000},  xticklabels={$0$,$10k$,$20k$,$30k$},  scaled x ticks=false,      height=0.17\textheight, width=0.485\textwidth,
ymin=-114.449675, ymax=125,
ytick style={color=black}
]
\path [draw=green, fill=green, opacity=0.1]
(axis cs:1000,-98.954954954955)
--(axis cs:1000,-103.317)
--(axis cs:2000,-51.284)
--(axis cs:3000,-33.49)
--(axis cs:4000,-17.638)
--(axis cs:5000,-4.8655)
--(axis cs:6000,9.734)
--(axis cs:7000,13.7815)
--(axis cs:8000,15.8405)
--(axis cs:9000,25.6465)
--(axis cs:10000,31.513)
--(axis cs:11000,41.1205)
--(axis cs:12000,46.4035)
--(axis cs:13000,50.2775)
--(axis cs:14000,47.977)
--(axis cs:15000,47.3825)
--(axis cs:16000,54.2055)
--(axis cs:17000,55.6495)
--(axis cs:18000,50.5055)
--(axis cs:19000,52.272)
--(axis cs:20000,51.469)
--(axis cs:21000,53.081)
--(axis cs:22000,57.4625)
--(axis cs:23000,56.109)
--(axis cs:24000,52.5545)
--(axis cs:25000,56.0245)
--(axis cs:26000,57.565)
--(axis cs:27000,50.089)
--(axis cs:28000,54.2835)
--(axis cs:29000,53.982)
--(axis cs:30000,55.743)
--(axis cs:30000,63.71)
--(axis cs:30000,63.71)
--(axis cs:29000,59.3175)
--(axis cs:28000,56.4655)
--(axis cs:27000,57.796)
--(axis cs:26000,59.3755)
--(axis cs:25000,59.0975)
--(axis cs:24000,58.413)
--(axis cs:23000,60.206)
--(axis cs:22000,60.3015)
--(axis cs:21000,62.85)
--(axis cs:20000,55.823)
--(axis cs:19000,55.3975)
--(axis cs:18000,57.906)
--(axis cs:17000,58.387)
--(axis cs:16000,56.324)
--(axis cs:15000,57.9015)
--(axis cs:14000,58.8625)
--(axis cs:13000,51.7615)
--(axis cs:12000,51.616)
--(axis cs:11000,41.321)
--(axis cs:10000,39.82)
--(axis cs:9000,34.269)
--(axis cs:8000,22.988)
--(axis cs:7000,18.7615)
--(axis cs:6000,15.0545)
--(axis cs:5000,-3.3115)
--(axis cs:4000,-13.2865)
--(axis cs:3000,-20.492)
--(axis cs:2000,-35.1855)
--(axis cs:1000,-98.954954954955)
--cycle;

\path [draw=red, fill=red, opacity=0.1]
(axis cs:1000,-98.1981981981982)
--(axis cs:1000,-105.96)
--(axis cs:2000,-46.092)
--(axis cs:3000,-28.3565)
--(axis cs:4000,-13.25)
--(axis cs:5000,-2.2815)
--(axis cs:6000,11.5465)
--(axis cs:7000,20.439)
--(axis cs:8000,19.179)
--(axis cs:9000,34.35)
--(axis cs:10000,40.521)
--(axis cs:11000,41.4045)
--(axis cs:12000,47.5965)
--(axis cs:13000,46.943)
--(axis cs:14000,56.5005)
--(axis cs:15000,54.691)
--(axis cs:16000,55.538)
--(axis cs:17000,56.891)
--(axis cs:18000,55.6365)
--(axis cs:19000,56.429)
--(axis cs:20000,57.648)
--(axis cs:21000,55.35)
--(axis cs:22000,53.313)
--(axis cs:23000,56.2825)
--(axis cs:24000,55.674)
--(axis cs:25000,52.53)
--(axis cs:26000,54.2155)
--(axis cs:27000,56.013)
--(axis cs:28000,58.2875)
--(axis cs:29000,57.92)
--(axis cs:30000,56.5735)
--(axis cs:30000,61.164)
--(axis cs:30000,61.164)
--(axis cs:29000,59.2725)
--(axis cs:28000,61.8505)
--(axis cs:27000,58.4975)
--(axis cs:26000,57.8925)
--(axis cs:25000,59.6325)
--(axis cs:24000,56.659)
--(axis cs:23000,62.725)
--(axis cs:22000,58.994)
--(axis cs:21000,60.9315)
--(axis cs:20000,62.9785)
--(axis cs:19000,60.719)
--(axis cs:18000,58.62)
--(axis cs:17000,60.8805)
--(axis cs:16000,59.7655)
--(axis cs:15000,63.8335)
--(axis cs:14000,61.008)
--(axis cs:13000,53.2245)
--(axis cs:12000,50.3775)
--(axis cs:11000,46.5645)
--(axis cs:10000,45.387)
--(axis cs:9000,39.348)
--(axis cs:8000,29.305)
--(axis cs:7000,23.0225)
--(axis cs:6000,14.5175)
--(axis cs:5000,8.865)
--(axis cs:4000,-11.2375)
--(axis cs:3000,-26.0695)
--(axis cs:2000,-43.02)
--(axis cs:1000,-98.1981981981982)
--cycle;

\addplot [semithick, green]
table {%
1000 -100.793
2000 -45.0775
3000 -26.2525
4000 -15.0553
5000 -3.8835
6000 11.6608
7000 15.8847
8000 18.4002
9000 28.7078
10000 34.9358
11000 41.242
12000 49.5455
13000 51.0492
14000 53.6215
15000 52.9795
16000 55.2723
17000 57.3132
18000 55.1578
19000 53.8525
20000 53.1907
21000 57.3147
22000 59.1637
23000 58.775
24000 55.7553
25000 57.479
26000 58.185
27000 53.8228
28000 55.0383
29000 56.311
30000 59.6288
};
\addplot [semithick, red]
table {%
1000 -101.876
2000 -44.2992
3000 -27.2957
4000 -12.2798
5000 3.473
6000 12.8887
7000 21.6958
8000 25.4168
9000 36.1323
10000 43.191
11000 44.7032
12000 49.4037
13000 50.3647
14000 58.374
15000 60.2673
16000 57.1103
17000 58.8648
18000 57.0805
19000 58.9053
20000 59.6227
21000 57.9783
22000 56.0683
23000 59.4975
24000 56.3273
25000 56.9055
26000 56.1163
27000 57.6278
28000 60.5065
29000 58.7793
30000 58.1347
};
\end{groupplot}

\end{tikzpicture}
    \caption{Return gained by intermediate policies throughout
    RL in the \emph{slippery shortcuts} gridworlds.
    The x-axes
    display the return and the y-axes display the episodes
    at which policies are evaluated.
    Green plots: shielded agent, red plots: unshielded agent.}
    \label{fig:gridnworld_2_results}
\end{figure}
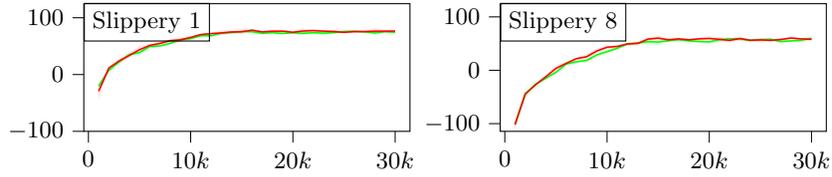
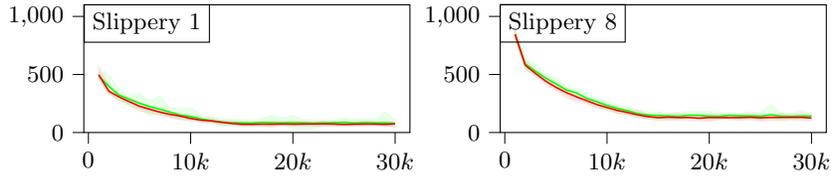
\begin{figure}[t]
    \centering
\begin{tikzpicture}

\definecolor{darkgray176}{RGB}{176,176,176}
\definecolor{lightgray204}{RGB}{204,204,204}
\pgfplotsset{every axis title/.append style={below right,at={(0,0.88), font=\small},draw=black}}

\begin{groupplot}[group style={group size=2 by 1,horizontal sep=1.2cm,vertical sep=0.6cm}]
\nextgroupplot[
tick align=outside,
tick pos=left,
title={Slippery 1},
x grid style={darkgray176},
xmin=-450, xmax=31450,
xtick style={color=black},
y grid style={darkgray176},
 xtick={0,10000,20000,30000},  xticklabels={$0$,$10k$,$20k$,$30k$},  scaled x ticks=false,      height=0.17\textheight, width=0.485\textwidth,
ymin=0, ymax=1100,
ytick style={color=black}
]
\path [draw=green, fill=green, opacity=0.1]
(axis cs:1000,530)
--(axis cs:1000,452)
--(axis cs:2000,333)
--(axis cs:3000,278)
--(axis cs:4000,228)
--(axis cs:5000,204)
--(axis cs:6000,143)
--(axis cs:7000,152)
--(axis cs:8000,135)
--(axis cs:9000,105)
--(axis cs:10000,110)
--(axis cs:11000,91)
--(axis cs:12000,82)
--(axis cs:13000,49)
--(axis cs:14000,55)
--(axis cs:15000,49)
--(axis cs:16000,49)
--(axis cs:17000,47)
--(axis cs:18000,58)
--(axis cs:19000,64)
--(axis cs:20000,35)
--(axis cs:21000,24)
--(axis cs:22000,32)
--(axis cs:23000,62)
--(axis cs:24000,54)
--(axis cs:25000,59)
--(axis cs:26000,59)
--(axis cs:27000,58)
--(axis cs:28000,62)
--(axis cs:29000,62)
--(axis cs:30000,36)
--(axis cs:30000,114)
--(axis cs:30000,114)
--(axis cs:29000,178)
--(axis cs:28000,118)
--(axis cs:27000,117)
--(axis cs:26000,114)
--(axis cs:25000,127)
--(axis cs:24000,108)
--(axis cs:23000,110)
--(axis cs:22000,110)
--(axis cs:21000,111)
--(axis cs:20000,146)
--(axis cs:19000,125)
--(axis cs:18000,150)
--(axis cs:17000,116)
--(axis cs:16000,106)
--(axis cs:15000,111)
--(axis cs:14000,112)
--(axis cs:13000,122)
--(axis cs:12000,132)
--(axis cs:11000,153)
--(axis cs:10000,216)
--(axis cs:9000,201)
--(axis cs:8000,211)
--(axis cs:7000,317)
--(axis cs:6000,269)
--(axis cs:5000,343)
--(axis cs:4000,317)
--(axis cs:3000,374)
--(axis cs:2000,495)
--(axis cs:1000,530)
--cycle;

\path [draw=red, fill=red, opacity=0.1]
(axis cs:1000,570)
--(axis cs:1000,462)
--(axis cs:2000,330)
--(axis cs:3000,278)
--(axis cs:4000,236)
--(axis cs:5000,187)
--(axis cs:6000,167)
--(axis cs:7000,156)
--(axis cs:8000,125)
--(axis cs:9000,115)
--(axis cs:10000,99)
--(axis cs:11000,90)
--(axis cs:12000,82)
--(axis cs:13000,67)
--(axis cs:14000,63)
--(axis cs:15000,58)
--(axis cs:16000,55)
--(axis cs:17000,55)
--(axis cs:18000,47)
--(axis cs:19000,60)
--(axis cs:20000,55)
--(axis cs:21000,55)
--(axis cs:22000,56)
--(axis cs:23000,54)
--(axis cs:24000,57)
--(axis cs:25000,58)
--(axis cs:26000,56)
--(axis cs:27000,59)
--(axis cs:28000,60)
--(axis cs:29000,53)
--(axis cs:30000,52)
--(axis cs:30000,93)
--(axis cs:30000,93)
--(axis cs:29000,92)
--(axis cs:28000,90)
--(axis cs:27000,95)
--(axis cs:26000,89)
--(axis cs:25000,91)
--(axis cs:24000,86)
--(axis cs:23000,99)
--(axis cs:22000,92)
--(axis cs:21000,95)
--(axis cs:20000,94)
--(axis cs:19000,94)
--(axis cs:18000,92)
--(axis cs:17000,89)
--(axis cs:16000,87)
--(axis cs:15000,90)
--(axis cs:14000,103)
--(axis cs:13000,104)
--(axis cs:12000,128)
--(axis cs:11000,125)
--(axis cs:10000,148)
--(axis cs:9000,160)
--(axis cs:8000,177)
--(axis cs:7000,195)
--(axis cs:6000,222)
--(axis cs:5000,262)
--(axis cs:4000,303)
--(axis cs:3000,331)
--(axis cs:2000,403)
--(axis cs:1000,570)
--cycle;

\addplot [semithick, green]
table {%
1000 488.808
2000 398.192
3000 321.654
4000 286.115
5000 252.115
6000 222.5
7000 201.5
8000 174.846
9000 149.615
10000 137.885
11000 117.5
12000 103.038
13000 90.9231
14000 84.0769
15000 83.5769
16000 81.8462
17000 87.8077
18000 85.0769
19000 83.4231
20000 85.2308
21000 83.5769
22000 80.9615
23000 82.6538
24000 85.4615
25000 87.4615
26000 82.0385
27000 85.4615
28000 84.1538
29000 83.1923
30000 79.1538
};
\addplot [semithick, red]
table {%
1000 498.077
2000 355.385
3000 306.308
4000 265.308
5000 225.923
6000 200.423
7000 175.308
8000 155.615
9000 141.385
10000 120.846
11000 106.577
12000 99.6538
13000 88.1538
14000 78
15000 70.6154
16000 70.3462
17000 71.9231
18000 71.3077
19000 73.3846
20000 71.6154
21000 72.9231
22000 73.5
23000 74.3462
24000 72.8846
25000 69.0769
26000 71.1538
27000 72.6538
28000 72.3077
29000 69.9231
30000 74.5385
};

\nextgroupplot[
tick align=outside,
tick pos=left,
title={Slippery 8},
title style={above left,at={(0.385,0.58)}},
x grid style={darkgray176},
xmin=-450, xmax=31450,
xtick style={color=black},
y grid style={darkgray176},
 xtick={0,10000,20000,30000},  xticklabels={$0$,$10k$,$20k$,$30k$},  scaled x ticks=false,      height=0.17\textheight, width=0.485\textwidth,
ymin=0, ymax=1100,
ytick style={color=black}
]
\path [draw=green, fill=green, opacity=0.1]
(axis cs:1000,899)
--(axis cs:1000,807)
--(axis cs:2000,541)
--(axis cs:3000,491)
--(axis cs:4000,429)
--(axis cs:5000,377)
--(axis cs:6000,320)
--(axis cs:7000,305)
--(axis cs:8000,256)
--(axis cs:9000,231)
--(axis cs:10000,191)
--(axis cs:11000,166)
--(axis cs:12000,155)
--(axis cs:13000,127)
--(axis cs:14000,121)
--(axis cs:15000,110)
--(axis cs:16000,109)
--(axis cs:17000,119)
--(axis cs:18000,126)
--(axis cs:19000,121)
--(axis cs:20000,110)
--(axis cs:21000,119)
--(axis cs:22000,130)
--(axis cs:23000,113)
--(axis cs:24000,124)
--(axis cs:25000,108)
--(axis cs:26000,115)
--(axis cs:27000,110)
--(axis cs:28000,112)
--(axis cs:29000,122)
--(axis cs:30000,113)
--(axis cs:30000,170)
--(axis cs:30000,170)
--(axis cs:29000,186)
--(axis cs:28000,164)
--(axis cs:27000,168)
--(axis cs:26000,243)
--(axis cs:25000,169)
--(axis cs:24000,178)
--(axis cs:23000,168)
--(axis cs:22000,171)
--(axis cs:21000,172)
--(axis cs:20000,184)
--(axis cs:19000,184)
--(axis cs:18000,180)
--(axis cs:17000,173)
--(axis cs:16000,170)
--(axis cs:15000,176)
--(axis cs:14000,178)
--(axis cs:13000,195)
--(axis cs:12000,224)
--(axis cs:11000,239)
--(axis cs:10000,272)
--(axis cs:9000,298)
--(axis cs:8000,342)
--(axis cs:7000,380)
--(axis cs:6000,396)
--(axis cs:5000,478)
--(axis cs:4000,512)
--(axis cs:3000,592)
--(axis cs:2000,644)
--(axis cs:1000,899)
--cycle;

\path [draw=red, fill=red, opacity=0.1]
(axis cs:1000,892)
--(axis cs:1000,808)
--(axis cs:2000,547)
--(axis cs:3000,484)
--(axis cs:4000,412)
--(axis cs:5000,363)
--(axis cs:6000,299)
--(axis cs:7000,268)
--(axis cs:8000,250)
--(axis cs:9000,208)
--(axis cs:10000,196)
--(axis cs:11000,166)
--(axis cs:12000,152)
--(axis cs:13000,130)
--(axis cs:14000,117)
--(axis cs:15000,105)
--(axis cs:16000,105)
--(axis cs:17000,108)
--(axis cs:18000,105)
--(axis cs:19000,100)
--(axis cs:20000,94)
--(axis cs:21000,113)
--(axis cs:22000,111)
--(axis cs:23000,106)
--(axis cs:24000,111)
--(axis cs:25000,99)
--(axis cs:26000,104)
--(axis cs:27000,104)
--(axis cs:28000,111)
--(axis cs:29000,103)
--(axis cs:30000,105)
--(axis cs:30000,146)
--(axis cs:30000,146)
--(axis cs:29000,159)
--(axis cs:28000,155)
--(axis cs:27000,163)
--(axis cs:26000,155)
--(axis cs:25000,151)
--(axis cs:24000,155)
--(axis cs:23000,156)
--(axis cs:22000,145)
--(axis cs:21000,148)
--(axis cs:20000,157)
--(axis cs:19000,143)
--(axis cs:18000,150)
--(axis cs:17000,164)
--(axis cs:16000,157)
--(axis cs:15000,148)
--(axis cs:14000,160)
--(axis cs:13000,189)
--(axis cs:12000,192)
--(axis cs:11000,223)
--(axis cs:10000,247)
--(axis cs:9000,272)
--(axis cs:8000,300)
--(axis cs:7000,333)
--(axis cs:6000,387)
--(axis cs:5000,430)
--(axis cs:4000,475)
--(axis cs:3000,540)
--(axis cs:2000,626)
--(axis cs:1000,892)
--cycle;

\addplot [semithick, green]
table {%
1000 846.423
2000 591.231
3000 526.885
4000 468.5
5000 419.231
6000 366.923
7000 342
8000 294.962
9000 265.154
10000 232.769
11000 208.269
12000 188.038
13000 162.808
14000 150.385
15000 146.038
16000 143.385
17000 138.846
18000 149.154
19000 149.115
20000 140.692
21000 139.269
22000 148.769
23000 143.115
24000 143.423
25000 140.154
26000 152
27000 141.538
28000 138.615
29000 141.923
30000 141.923
};
\addplot [semithick, red]
table {%
1000 846.538
2000 580.038
3000 508.577
4000 442.654
5000 389.5
6000 344.577
7000 307.808
8000 274.462
9000 240.885
10000 215
11000 190.692
12000 172.231
13000 152
14000 136.962
15000 127.692
16000 131.846
17000 128.346
18000 129
19000 123.385
20000 129.154
21000 128.5
22000 127.538
23000 128.462
24000 131.308
25000 126.538
26000 129.962
27000 129.769
28000 130.769
29000 130.769
30000 124.962
};
\end{groupplot}

\end{tikzpicture}
    \caption{The number of safety violations throughout RL in the \emph{slippery shortcuts} gridworlds. The x-axes display the number of violations and y-axes display the episodes at which the policies are evaluated. Green plots: shielded agent, red plots: unshielded agent.}
    \label{fig:pitfrequency_2_results}
\end{figure}
\subsection{Slippery Shortcuts Gridworlds}
Next, we examine the performance of shielded and unshielded RL in the
\emph{slippery shortcuts} gridworlds illustrated in Figure~\ref{fig:slipperygw}.
Figure~\ref{fig:gridnworld_2_results} shows the average returns,
and Figure~\ref{fig:pitfrequency_2_results} shows the number of safety violations
gained by RL agents throughout learning. In contrast to the \emph{zigzag}
gridworlds, there is hardly any difference between the shielded and the
unshielded configurations for any size of the environment, neither for the return nor for the number of safety violations.
Therefore, we only printed the instances of \emph{slippery 1} and \emph{slippery 8} to safe space.
Moreover, there is very little variability, as the minimum and maximum returns (safety violations)
are very close to their average. Hence, performance is mostly
governed by RL and shielding has little influence. In these environments,
the agents succeed in finding safe paths without requiring assistance from
a shield. This may result from the pits being farther away from optimal paths,
as compared to the \emph{zigzag} gridworlds.

\begin{figure}[t]
    \centering
    \input{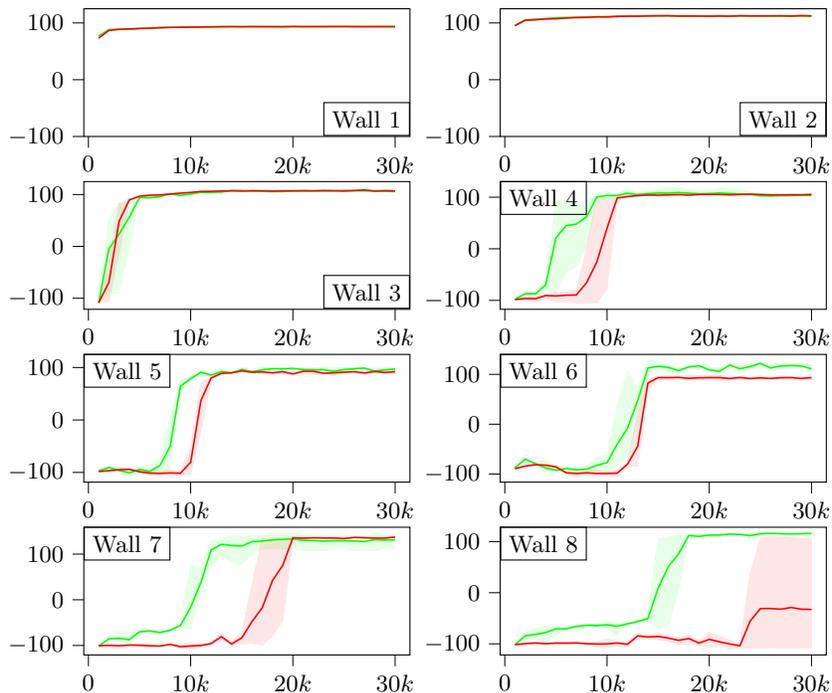}
    \caption{Return gained by intermediate policies throughout
    RL in the \emph{walls} gridworlds.
    The x-axes
    display the return and the y-axes display the episodes
    at which policies are evaluated.
    Green plots: shielded agent, red plots: unshielded agent.}
    \label{fig:gridnworld_3_results}
\end{figure}
\begin{figure}[h]
    \centering
    \input{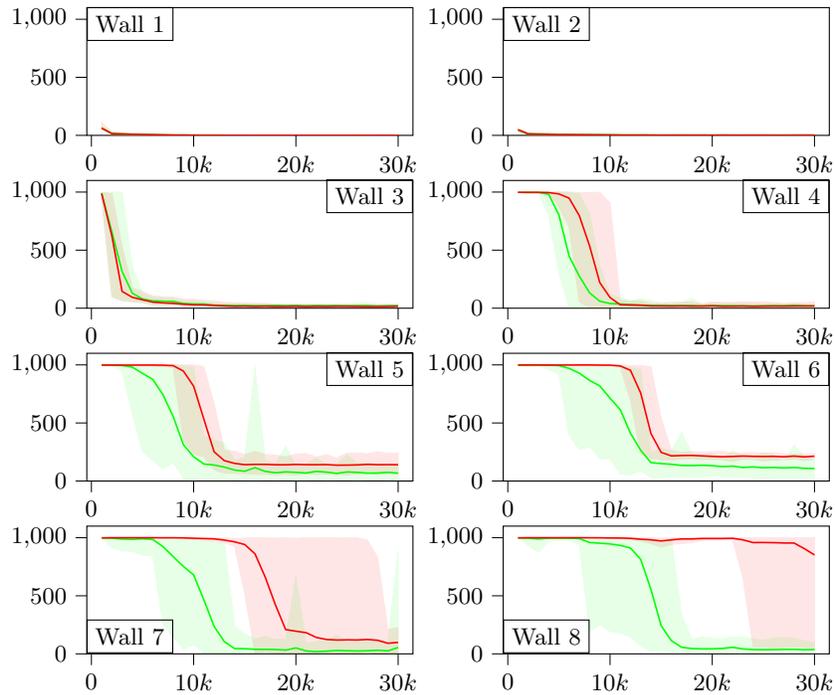}
    \caption{The number of safety violations throughout RL in the \emph{walls} gridworlds. The x-axes display the number of violations and y-axes display the episodes at which the policies are evaluated. Green plots: shielded agent, red plots: unshielded agent.}
    \label{fig:pitfrequency_3_results}
\end{figure}
\subsection{Wall Gridworlds}
In the following, we discuss the experiments performed on
the \emph{walls} gridworlds of Figure~\ref{fig:wallgw}. Figure~\ref{fig:gridnworld_3_results}
shows the returns,
and Figure~\ref{fig:pitfrequency_3_results} shows the number of safety violations gained at different stages of learning. As
for the \emph{zigzag} gridworlds, we see hardly any difference
between shielded and unshielded RL for the
three smallest environments, whereas shielded RL performs better in the
larger environments. Unlike before, however, there is less variability
in the performance and number of safety violations of shielded RL. In Figure~\ref{fig:wallgw}, we can see
that the slippery tiles are farther away from the optimal route, which
is also true for the larger \emph{walls} gridworlds.
As a result, learned MDPs do not need to be as accurate with respect
to probability estimations for effective shields
to be created. Especially for the largest example, \emph{Wall 8}, shielding
improves performance and reduces the number of safety violations considerably. It takes about $18000$ episodes to
learn a policy that consistently reaches the goal in all $30$ repetitions
of the corresponding experiments. In contrast, unshielded RL fails to consistently
find a good policy even after $30000$ episodes.

\begin{figure}[t]
    \centering
    \input{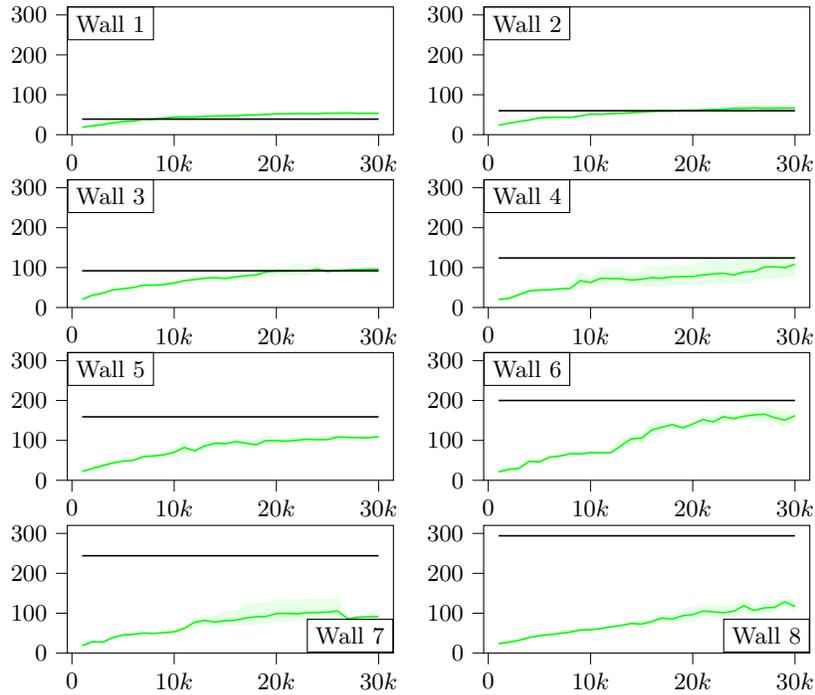}
    \caption{Average size of learned MDPs for the \emph{walls} gridworlds
    (x-axes) plotted in green compared to the size of the environment
    plotted as a black line. The y-axes display the episode at which
    the MDP size was measured.}
    \label{fig:gridnworld_2_automata_sizes}
\end{figure}

Finally, let us investigate a potential reason for performance improvements
resulting from shielding or the absence thereof. In Figure~\ref{fig:gridnworld_2_automata_sizes}, we show the size of learned
MDPs compared to the size of the \emph{walls} environments, i.e., the number of
tiles in every environment. Since the environment size is constant
in an experiment, it is shown as a black straight line.
The learned MDP size, measured in the
number of states, generally increases throughout the learning
due to more information getting available.
It can be seen that for the first three environment
sizes, the final learned MDPs are slightly larger than the
environment. Hence, these MDPs cannot represent the environments
and their safety-relevant features more efficiently than a Q-table.
The fact that learned MDPs are even larger
than the environments from which
they are learned results from two properties of our learning setup.
First, MDPs learned by \IOALERGIA only converge in the large
sample limit to the true underlying MDPs~\cite{DBLP:journals/ml/MaoCJNLN16}.
There are no guarantees for MDPs learned from finite amounts of data.
Second and more importantly, abstraction introduces non-determinism, while
MDP learning basically performs a determinization of the resulting
non-deterministic MDP. This determinization causes the number of states
to increase, similar to the construction of belief MDPs from partially
observable MDPs~\cite{DBLP:conf/aaai/CassandraKL94}.

When the environment is larger than the learned MDP modeling safety-relevant features
of the environment, shielding improves RL performance.
This holds for all environments from \emph{Wall 4} throughout \emph{Wall 8}.
Comparing Figure~\ref{fig:gridnworld_3_results} or Figure~\ref{fig:pitfrequency_3_results} with Figure~\ref{fig:gridnworld_2_automata_sizes},
we can see that the larger the size difference between the environment and the learned MDP is,
the larger the performance impact or reduction in the number of safety violations.

\subsection{Discussion}
We conclude this section with a discussion of the main results of our
experiments and some insights that we gained. Our results show
that learned shields can improve RL performance
as illustrated by our first and third set of experiments.
In the first case
of the \emph{zigzag} gridworlds,
the agent has to traverse along tiles located closely to pits in order to reach the goal. Therefore, a shield is able to prevent many safety violations.
In the case of the \emph{walls} gridworlds, we observed
that learning shields especially pays off when the learned safety MDP
is much smaller than the complete environment.
We will explore this connection in future work, as it may enable scaling
to larger environments. Deep reinforcement learning
can efficiently solve tasks in complex environments, such as, computer games~\cite{DBLP:journals/corr/MnihKSGAWR13}, while we can abstract away
non-safety-relevant details to learn small MDPs for shielding.

In the \emph{slippery shortcut} gridworlds, we observed that shielding
does not necessarily improve performance. It seems easier for the
agent to infer through RL how to navigate safely than in the other examples.

\section{Conclusion}
\label{sec:conclusion}
We presented an approach for iterative safe reinforcement learning via learned shields.
At runtime, we learn environmental models from collected traces and continuously update shields
that prevent safety violations during execution. 
RL with learned shields comprises three steps:
(1) An RL agent exploring the environment, protected by a shield,
and collecting abstracted experiences, which
represent safety-information about the environment.
(2) Learning a deterministic 
labeled MDP from the collected data of the RL agent. (3)
Synthesizing a shield from such an MDP.

In contrast to most previous work on shielding, which
commonly requires abstract environment models, the proposed
approach is model-free and therefore applicable in black-box environments.
We learn environment models solely from experiences of the
RL agent. The agent can also infer 
its policy using a model-free approach, such as, Q-learning~\cite{DBLP:journals/ml/WatkinsD92}. 
The downside is that we cannot enforce absolute safety. In order
to learn safety-relevant information, the agent needs to experience
some safety violations. Despite this limitation, our evaluation 
shows that in most cases, 
RL with learned shields converges more quickly than unshielded RL.
Since optimal policies in our experiments should inflict hardly any
safety violations, faster convergence implies that shielded agents run 
into fewer safety violations. 

In future work, we will explore RL with learned shields in environments
of larger size, where we aim to combine deep RL with MDP learning. 
Our intuition is that we can generally represent safety-relevant environmental features concisely with an MDP over abstract observations. We expect this to be true even if, 
for instance, the agent perceives its environment by processing
high-resolution images. 
In addition to other (deep) RL techniques, we will explore different automata learning
techniques. For instance, we could integrate RL more directly into active automata learning
of stochastic system models~\cite{DBLP:journals/fac/TapplerA0EL21,DBLP:conf/sefm/TapplerMAP21}. By learning timed automata~\cite{DBLP:conf/nfm/MediouniNBB17,DBLP:conf/icgi/VerwerWW10,DBLP:conf/formats/TapplerALL19,DBLP:conf/nfm/AichernigPT20}, we could extend learning-based shielding to systems with time-dependent behaviour.

\subsubsection*{Acknowledgments.}
This work has been supported by the "University SAL Labs" initiative of Silicon Austria Labs (SAL) and its Austrian partner universities for applied fundamental research for electronic based systems.
Additionally, this project has received funding from the European Union’s Horizon 2020 research
and innovation programme under grant agreement N◦ 956123 - FOCETA.
%

\bibliographystyle{splncs04}
\bibliography{bibliography}

\end{document}